\documentclass[10pt,twoside,twocolumn,english]{IEEEtran}
\usepackage{mathptmx}

\usepackage[T1]{fontenc}
\usepackage[latin9]{inputenc}
\usepackage{color}
\usepackage{array}
\usepackage{float}
\usepackage{booktabs}
\usepackage{textcomp}
\usepackage{multirow}
\usepackage{amsmath}
\usepackage{amssymb}
\usepackage{stmaryrd}
\usepackage{stackrel}
\usepackage{graphicx}
\usepackage{epstopdf}

\makeatletter

\providecommand{\tabularnewline}{\\}
\floatstyle{ruled}
\newfloat{algorithm}{tbp}{loa}
\providecommand{\algorithmname}{Algorithm}
\floatname{algorithm}{\protect\algorithmname}

\newenvironment{lyxcode}
	{\par\begin{list}{}{
		\setlength{\rightmargin}{\leftmargin}
		\setlength{\listparindent}{0pt}
		\raggedright
		\setlength{\itemsep}{0pt}
		\setlength{\parsep}{0pt}
		\normalfont\ttfamily}%
	 \item[]}
	{\end{list}}

\usepackage{algorithm}
\usepackage{algpseudocode}
\usepackage{flushend}

\usepackage[numbers,sort&compress]{natbib} 

\usepackage{colortbl}
\definecolor{grey}{cmyk}{0, 0, 0, 0.4}

\usepackage{flafter}

\makeatother

\usepackage{babel}
\begin{document}
\title{An Online Self-learning Graph-based Lateral Controller for Self-Driving
Cars}
\author{Jilan Samiuddin, Benoit Boulet and Di Wu}
\IEEEaftertitletext{\noindent \thanks{Manuscript received ; revised ; accepted
. This work was supported by Quebec\textquoteright s Fonds de
recherche Nature et technologies. (Corresponding authors: Jilan Samiuddin;
Di Wu.)\\The authors are with the Intelligent Automation Lab of the
Centre for Intelligent Machines and the Department of Electrical and
Computer Engineering, McGill University, Montreal, QC H3A 0E9, Canada
(e-mail: jilan. samiuddin@mail.mcgill.ca; benoit.boulet@mcgill.ca;
di.wu5@mcgill.ca).}}

\maketitle
\begin{abstract}
The hype around self-driving cars has been growing over the past years
and has sparked much research. Several modules in self-driving cars
are thoroughly investigated to ensure safety, comfort, and efficiency,
among which the controller is crucial. The controller module can be
categorized into longitudinal and lateral controllers in which the
task of the former is to follow the reference velocity, and the latter
is to reduce the lateral displacement error from the reference path.
Generally, a tuned controller is not sufficient to perform in all
environments. Thus, a controller that can adapt to changing conditions
is necessary for autonomous driving. Furthermore, these controllers
often depend on vehicle models that also need to adapt over time due
to varying environments. This paper uses graphs to present novel techniques
to learn the vehicle model and the lateral controller online. First,
a heterogeneous graph is presented depicting the current states of
and inputs to the vehicle. The vehicle model is then learned online
using known physical constraints in conjunction with the processing
of the graph through a Graph Neural Network structure. Next, another
heterogeneous graph -- depicting the transition from current to desired
states -- is processed through another Graph Neural Network structure
to generate the steering command on the fly. Finally, the performance
of this self-learning model-based lateral controller is evaluated
and shown to be satisfactory on an open-source autonomous driving
platform called CARLA.
\end{abstract}


\section*{\textcolor{black}{Nomenclature}}
\begin{flushleft}
{\small{}}%
\begin{tabular}{c>{\raggedright}p{2.5in}}
\textit{\small{}Variables} & \tabularnewline
{\small{}$s,d$} & {\small{}Longitudinal, lateral positions (Frenet coordinates)}\tabularnewline
{\small{}$x,y$} & {\small{}Cartesian coordinates}\tabularnewline
{\small{}$\theta$} & {\small{}Heading in radians}\tabularnewline
{\small{}$v$} & {\small{}Velocity in m/s}\tabularnewline
{\small{}$h,Z$} & {\small{}Feature or node embedding array}\tabularnewline
{\small{}$E$} & {\small{}Edge connection array of a graph}\tabularnewline
{\small{}$X$} & {\small{}States of the ego}\tabularnewline
{\small{}$Y$} & {\small{}Output of the ego}\tabularnewline
{\small{}$\alpha$} & {\small{}Attention coefficient in GAT network}\tabularnewline
{\small{}$\delta$} & {\small{}Steering command}\tabularnewline
{\small{}$T$} & {\small{}Throttle command}\tabularnewline
\end{tabular}{\small\par}
\par\end{flushleft}

\begin{flushleft}
{\small{}}%
\begin{tabular}{c>{\raggedright}p{2.5in}}
{\small{}$e$} & {\small{}Error signal}\tabularnewline
 & \tabularnewline
\textit{\small{}Parameters} & \tabularnewline
{\small{}$W$} & {\small{}Learnable weights }\tabularnewline
{\small{}$B$} & {\small{}Learnable biases}\tabularnewline
 & \tabularnewline
\textit{\small{}Symbols} & \tabularnewline
{\small{}$\mathcal{N}$} & {\small{}Neighborhood of a node}\tabularnewline
{\small{}$N$} & {\small{}Number of nodes}\tabularnewline
{\small{}$G$} & {\small{}Graph}\tabularnewline
{\small{}$R$} & {\small{}Reference path coordinates array}\tabularnewline
{\small{}$M$} & {\small{}Output dimension of G2O}\tabularnewline
 & \tabularnewline
\multicolumn{2}{l}{\textit{\small{}Symbols (Superscript)}}\tabularnewline
{\small{}$w$} & {\small{}Wheel of ego}\tabularnewline
{\small{}$r,f$} & {\small{}Rear end and front end of ego}\tabularnewline
 & \tabularnewline
\multicolumn{2}{l}{\textit{\small{}Symbols (Subscript)}}\tabularnewline
{\small{}$r,l$} & {\small{}Right and left side of ego}\tabularnewline
{\small{}$c$} & {\small{}Vehicle controller}\tabularnewline
{\small{}$m$} & {\small{}Vehicle model}\tabularnewline
\end{tabular}{\small\par}
\par\end{flushleft}

\section{Introduction}

The Automated Driving Systems (ADS) industry has experienced a surge
in investment and research -- the focus of the researchers are primarily,
but not limited to, safety, efficiency and convenience \cite{cao2022future}.
A study has estimated that 90\% of fatalities in motor vehicle accidents
are driven by human errors in contrast to only 2\% resulting from
a vehicle malfunction \cite{singh2015critical}. Promising solutions
in self-driving cars are emerging to elevate safety and reshape the
transportation industry \cite{joerger2017towards}. 

One of the significant tasks of the self-driving car, also referred
to as the ego, is to drive along a predefined trajectory. This requires
solving the control problems for longitudinal and lateral dynamics
of the vehicle \cite{jiang2018lateral}. This paper primarily focuses
on solving the control problem of the lateral dynamics only. Lateral
controllers using fuzzy control \cite{yang2007expert}, $H_{\infty}$
control \cite{huang2013robust}, sliding mode control \cite{lee2017predictive}
have been studied. Another approach is the geometric control, such
as the Pure Pursuit controller \cite{coulter1992implementation} and
the Stanley controller \cite{thrun2006stanley} that utilizes the
geometry of the vehicle kinematics and the reference trajectory to
compute the steering commands. Performance comparison of these controllers
is presented in \cite{dominguez2016comparison}. However, these methods
rely on linear models of the lateral dynamics. In real-world driving
scenarios, there are lots of uncertainties in the environments and
one particular challenge in ADS is how to deal with varying environments
(e.g., road networks, weather conditions, etc.) \cite{muhammad2020deep}.
For classical solutions, for example, the control of these cars was
mostly rule-based and required painstaking manual tuning of the control
parameters. Yet, such control laws were challenging to generalize
to individual scenarios \cite{kuutti2020survey}. Thus, it is necessary
that autonomous cars can learn on the fly, i.e., they adopt online
learning strategies. 

Deep learning allows control agents to adapt to dynamic environments
and generalize to new settings using self-optimization through time.
Control of the ego car, in general, can be divided into two categories
\cite{li2019reinforcement}: (1) end-to-end, and (2) the perception
and control separation methods. The end-to-end learning controllers
directly process sensor readings to control desired output variables.
In contrast, the perception and control separation technique has a
separate perception module that extracts the features (e.g., position,
velocity, etc.) and the features are then used to make decisions by
the controller module. In an early end-to-end method, \cite{pomerleau1997neural}
used a simple feedforward network to train with camera images to generate
steering commands. Convolutional Neural Networks (CNN) have also been
a popular choice to process camera images for lateral control of the
vehicle \cite{chen2017end,rausch2017learning}. Since end-to-end learning
methods typically predict a control input for a particular observation
and the predicted action affects the following observation, the error
can accumulate and lead the network to a completely different observation
\cite{ross2010efficient}. Another drawback of end-to-end learning
is that it requires a large dataset for training \cite{ross2011reduction}.
Kwon et al. \cite{kwon2022incremental} proposed an incremental end-to-end
learning method where the training is initiated with data collected
by a human driver and later replaced by the neural network for further
learning. \textcolor{black}{Khalil et al. \cite{khalil2023anec} introduced
an adaptive neural lateral controller for end-to-end learning in which
they utilize a base model and a predictive model inspired by human's
near/far gaze distribution. Despite promising results achieved by
end-to-end learning techniques, these methods suffer from poor interpretability
and generlization to new scenes \cite{zhu2023learning}. }In our work,
we use the perception and control separation technique; however, we
assume perfect state measurements, since the perception module is
out of the scope of this work.

Reinforcement Learning (RL) is another frequent choice for lateral
control. Ma et al. \cite{ma2024game} propose a game-theoretic receding
horizon reinforcement learning lateral controller by formulating the
uncertainties on the ego as a player by zero-sum differential games.
In addition, an actor-critic algorithm combined with a critic neural
network and two actor neural networks define their control startegy.
Brasch et al. \cite{brasch2022lateral} used the soft-actor-critic
algorithm of RL to determine steering values to reduce path tracking
error. Wasala et al. \cite{wasala2020trajectory} proposed a novel
reward function to minimize tracking errors and improve comfort and
safety for their RL algorithm. For lane changing maneuvers, Wang et
al. \cite{wang2018reinforcement} proposed a quadratic function as
the Q-function for their RL algorithm and use neural networks to approximate
the coefficients of the function. While the RL has the advantage of
not requiring a vehicle model, it essentially trains in a trial-and-error
mode, making it dangerous to train an agent on an actual vehicle.
Our proposed methods allow safer training of the actual vehicle in
real time. 

Another potential method for lateral control is the Model Predictive
Control (MPC) that has demonstrated reasonable control performance
\cite{shen2017mpc}\cite{falcone2007linear}. Costa et al. \cite{costa2023online}
implemented a learning-based MPC algorithm to improve modeling accuracy
and perform online tuning of MPC parameters for control. While the
approach in \cite{costa2023online} is similar to ours, MPC is computationally
expensive and on-board computers of autonomous cars may not be powerful
enough to solve real-time optimization problems \cite{Zhou2023}.
In \cite{Zhou2023}, Zhou et al. mitigated the computational load
problem using event-triggered MPC in contrast to the generally used
time-triggered MPC; however, there is a slight performance loss. \textcolor{black}{To
achieve the task of platooning, Kazemi et al. \cite{kazemi2024longitudinal}
used a Laguerre-based MPC and a robust MPC for longitudinal and lateral
control, respectively. In \cite{jia2024rl}, RL was used to dynamically
update the objective weights of a nonlinear MPC to perform real-time
lateral control of the ego.}

\textcolor{black}{In recent efforts to enhance computational efficiency
in lateral controllers for autonomous vehicles, researchers have addressed
key implementation challenges. For instance, \cite{khosravian2024robust}
presents an integrated path planning and path-following system tailored
for real-world driving conditions. This system is optimized for computation
power limitations and validated through real-time inference on an
NVIDIA Jetson Nano within a Hardware-in-the-Loop (HIL) framework.
Another significant contribution is found in \cite{tork2021adaptive},
where computational costs are minimized by fixing some of the parameters
to -1, 0, and 1, reducing the complexity of the control algorithm.
While a comprehensive computational efficiency report is desirable,
this is out of the scope of this work and we report only the average
inference and training times of our proposed controller.}

\textcolor{black}{The main contributions of this paper are as follows:}
\begin{itemize}
\item Vehicle modeling: The vehicle is modeled using existing knowledge
of vehicle dynamics in unification with Graph Neural Network (GNN),
particularly Graph Attention Network, that processes a heterogeneous
graph depicting the vehicle. The strength of the model lies in the
fact that it can be trained online anytime the performance degrades. 
\item Lateral controller: A heterogeneous graph representing the state transition
of the vehicle -- from current to desired -- is processed through
a GNN-based network to generate steering commands. The controller
is competent in online learning, allowing it to adjust its parameters
and respond accordingly to unseen environments.
\item We have validated our methods in CARLA, an open-source high-fidelity
ADS platform \cite{dosovitskiy2017carla}. 
\end{itemize}
To the best of our knowledge, no literature exists that utilizes graphs
to solve lateral control problems in autonomous driving.

The rest of this article is organized as follows. Section \ref{sec:Preliminaries}
provides brief descriptions of the theoretical framework used in this
work. The proposed vehicle modeling and the lateral controller are
introduced in Section \ref{sec:Vehicle-Model-GLC}. In Section \ref{sec:Training-Process},
we discuss the procedure for training the vehicle model and the controller.
Section \ref{sec:Simulation-Results} presents the main simulation
results and a comparison with the baseline. Finally, Section \ref{sec:Conclusion}
concludes the article.

\section{Preliminaries\label{sec:Preliminaries}}

\subsection{Graph Neural Network\label{subsec:Graph-Neural-Network}}

\begin{figure*}[t]
\begin{centering}
\includegraphics[scale=0.2]{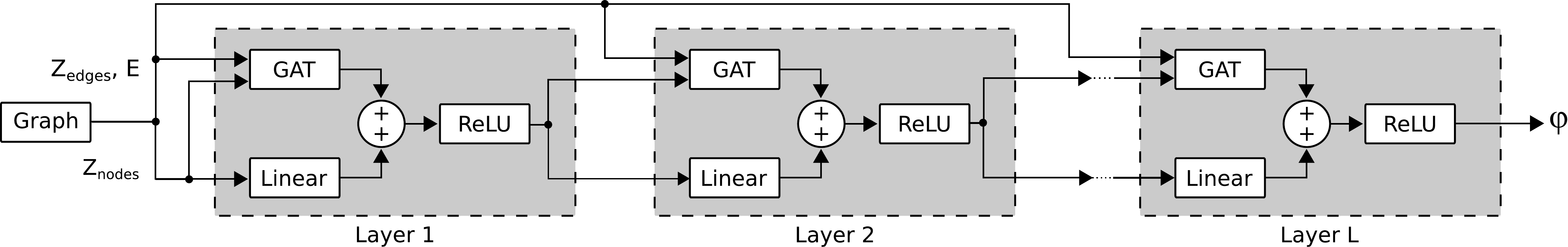}
\par\end{centering}
\begin{centering}
\vspace*{-1mm} 
\par\end{centering}
\caption{\textcolor{black}{Network architecture (graph-to-output, G2O) used
in this work to map a graph -- defined by its features $Z$ and edge
connections $E$ -- to output $\varphi$ \label{fig:Network-GAT}}}
\end{figure*}

Graphs exist all around us -- real world problems can be described
as connections between objects and, thus, as graphs. The objects in
a graph are termed as nodes and the connections between these objects
are called edges. While nodes have features, edges may or may not
have features associated with them. \textcolor{black}{Neural networks
that operate on graphs are called GNNs \cite{scarselli2008graph}.
For example, in node representations, the GNN learns a function $f$
that maps nodes of a homogeneou}s graph to a matrix $\mathbb{R}^{n\times m}$,
where, $n$ is the number of nodes and $m$ is the dimensionality
of the features of the nodes. For node $p$, it does so by aggregating
information from its neighbors (simplest being averaging messages
from its neighbors) and then applying a neural network in multiple
layers \cite{stanfordGNN}:{\small{}
\begin{equation}
\begin{aligned}h_{p}^{(i+1)} & =\sigma\left(W_{i}\underset{q\in\mathcal{N}(p)}{\sum}\frac{h_{q}^{(i)}}{\left|\mathcal{N}(p)\right|}+B_{i}h_{p}^{(i)}\right),\\
 & \forall i\in\left\{ 0,1,\ldots,L-1\right\} 
\end{aligned}
\end{equation}
}where, $h$ is the embedding of a node, $\sigma$ is a non-linear
activation function, $W_{i}$ and $B_{i}$ are the weights and biases
of the $i^{\mathrm{th}}$ layer respectively, $\mathcal{N}(p)$ is
the neighborhood of the target node $p$, and $L$ is the total number
of layers.

Graph Attention Network (GAT) \cite{velivckovic2017graph} was proposed
to integrate attention in GNNs so that the learning can be directed
to focus on more relevant segments of the input. Through the attention
mechanism, the network learns the importance of neighbors of a node
as it aggregates information from them. This importance, called the
attention coefficient ($e_{pq}$) between the target node $p$ and
its neighbor $q$, is computed by first applying a common linear transformation
$\boldsymbol{W}$ to the features ($\boldsymbol{h}$) of both $p$
and $q$, followed by applying a shared attentional mechanism $a$
as follows:{\small{}
\begin{equation}
e_{pq}=a\left(\boldsymbol{W}\boldsymbol{h}_{p},\boldsymbol{W}\boldsymbol{h}_{q}\right).
\end{equation}
}In \cite{velivckovic2017graph}, a single-layer feedforward neural
network is used for $a$. For the neighboring nodes to be comparable,
$e_{pq}$ is normalized using the softmax function:{\small{}
\begin{equation}
\alpha_{pq}=\frac{\mathrm{exp}\left(e_{pq}\right)}{\sum_{k\in\mathcal{N}(p)}\mathrm{exp}\left(e_{pk}\right)}.
\end{equation}
}Using the attention coefficients to weigh the importance of the neighboring
nodes, the final output features of every node is obtained as follows:{\small{}
\begin{equation}
\bar{h}_{p}=\sigma\left(\underset{q\in\mathcal{N}(p)}{\sum}\alpha_{pq}\boldsymbol{W}h_{q}\right),\label{eq:attention_mechanism}
\end{equation}
}where, $\sigma$ is a nonlinear activation function. To ensure stability
of the learning process of self-attention, \cite{velivckovic2017graph}
applies multi-head attention, i.e., $K$ independent attention mechanisms
of equation (\ref{eq:attention_mechanism}) are executed and then
their features are either concatenated or averaged. If concatenation
is applied, then the equation for the final output features of every
node is as follows:{\small{}
\begin{equation}
\bar{h}_{p}=\stackrel[k=1]{K}{\bigparallel}\sigma\left(\underset{q\in\mathcal{N}(p)}{\sum}\alpha_{pq}^{k}\boldsymbol{W}^{k}h_{q}\right),
\end{equation}
}where, $\bigparallel$ represents concatenation, $\alpha_{pq}^{k}$
are normalized attention coefficients computed by the $k^{\mathrm{th}}$
attention mechanism ($\mathrm{att}^{k}$), and $\boldsymbol{W}^{k}$
is the corresponding input linear transformation's weight matrix. 

In this work, the network architecture shown in Figure \ref{fig:Network-GAT}
is implemented to process a graph. The framework consists of multiple
layers, and each layer comprises a GAT and a linear layer in parallel,
the output of which is passed through a ReLU function. For convenience,
we will refer to this framework as G2O (graph-to-output) in the rest
of the paper.

\textcolor{black}{For Layer 1, the input is the original graph $G\left(Z,E\right)$,
where $Z$ is a collection of both node features $Z_{\mathrm{nodes}}$
and edge features $Z_{\mathrm{edges}}$. The output of the network
$\varphi$ is given by the following:}{\small{}
\begin{equation}
\begin{aligned}\bar{Z}_{1}= & \mathrm{ReLU}\left(\mathrm{GAT}^{(1)}\left(Z_{\mathrm{nodes}},Z_{\mathrm{edges}},E\right)+\right.\\
 & \left.\mathrm{Linear}^{(1)}\left(\left(Z_{\mathrm{nodes}}\right)^{f}\right)\right)\\
\bar{Z}_{l}= & \mathrm{ReLU}\left(\mathrm{GAT}^{(l)}\left(\bar{Z}_{l-1},Z_{\mathrm{edges}},E\right)+\right.\\
 & \left.\mathrm{Linear}^{(l)}\left(\left(\bar{Z}_{l-1}\right)^{f}\right)\right),\:\forall l=2,\ldots,L\\
\varphi= & \bar{Z}_{L}\in\mathbb{R}^{N\times M}
\end{aligned}
,
\end{equation}
}\textcolor{black}{where, $\left(\cdot\right)^{f}$ operator is the
flattening function \cite{Flatten_pytorch} that returns an array
collapsed into one dimension, $N$ is the number of nodes and $M$
is the output feature dimension of G2O.}
\begin{figure}[H]
\begin{centering}
\includegraphics[scale=0.1]{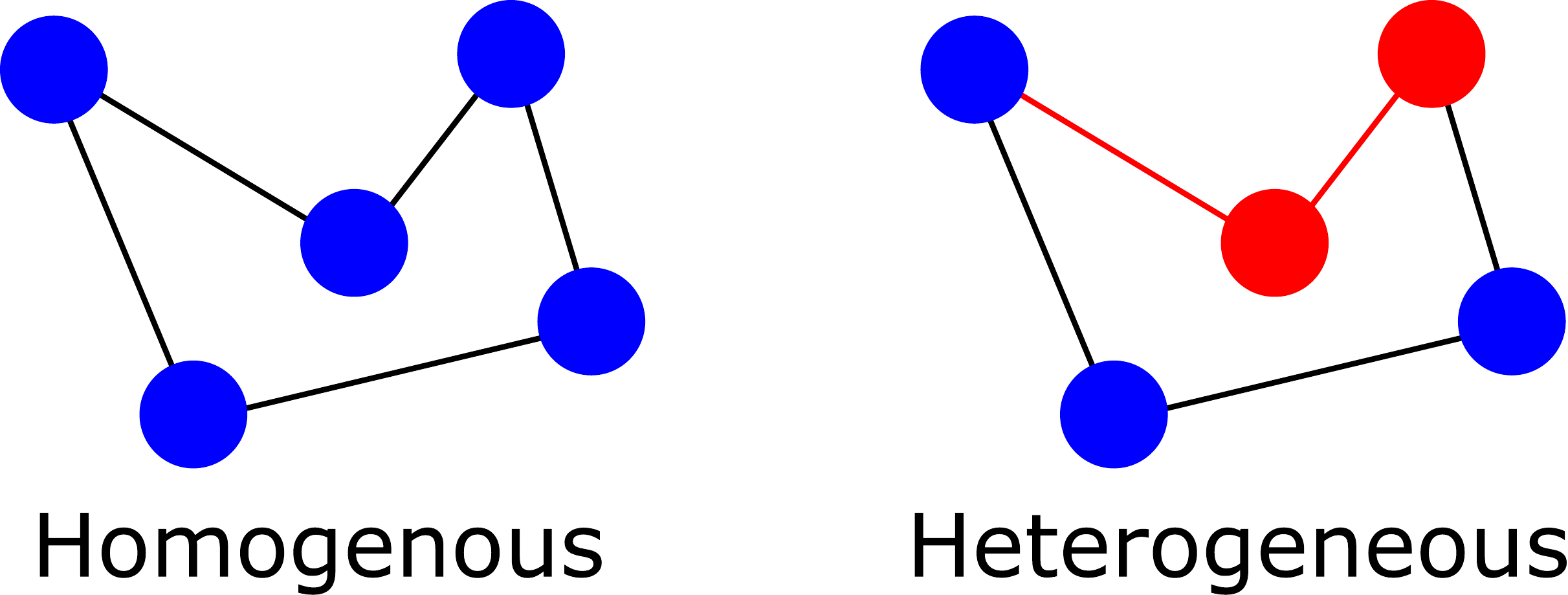}
\par\end{centering}
\begin{centering}
\vspace*{-5mm} 
\par\end{centering}
\textcolor{black}{\caption{\textcolor{black}{The homogeneous graph has only one type of node
(blue), and one type of edge (black), whereas, the heterogeneous graph
has two types of nodes (blue and red), and two types of edges (black
and red)}.\label{fig:homogeneous-heterogeneous}}
}
\end{figure}

\subsection{Heterogeneous Graphs}

Heterogeneous graphs, unlike homogeneous graphs, have nodes and/or
edges that can have various types or labels associated with them,
indicating their different roles or semantics. \textcolor{black}{Figure
\ref{fig:homogeneous-heterogeneous} shows an example of the difference
between the two types of graphs. }Because of the differences in type
and dimensionality, a single node or edge feature tensor is unable
to accommodate all the node or edge features of the graph \cite{Heterogeneous_PyG}.
In our work, we represent the wheels of the vehicles as well as its
actuators as nodes in section \ref{subsec:Proposed-Modeling}. However,
wheels are inherently different from the actuators, and thus, cannot
have the same feature representation. Similarly, when representing
transition from current state to desired state using a graph in section
\ref{sec:Controller-Architecture}, it is redundant to include the
velocity feature in the desired nodes because the purpose of the lateral
controller is not to track the longitudinal velocity. Thus, feature
dimensionality will be different between the current nodes and the
desired nodes, requiring a heterogeneous graph representation. Consequently,
the computation of message and update functions is conditioned on
node or edge type. For more details on how heterogeneous graphs are
processed, interested readers can refer to \cite{Heterogeneous_PyG}.
\begin{figure}[H]
\begin{centering}
\includegraphics[scale=0.35]{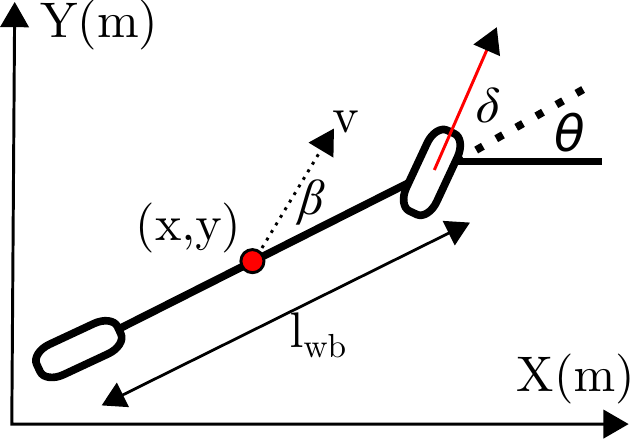}
\par\end{centering}
\begin{centering}
\vspace*{-1mm} 
\par\end{centering}
\caption{Dynamic vehicle bicycle model where the two axles are represented
as single wheels \label{fig:Dynamic-vehicle-bicycle}}
\end{figure}

\section{Vehicle Model and Lateral Controller\label{sec:Vehicle-Model-GLC}}

In this section, we propose a novel technique for the lateral controller
and an enhancement for the modeling of the vehicle by taking advantage
of the expressive power of graphs to represent real-world problems
(e.g. \cite{zheng2023interaction,li2020hierarchical}).

\subsection{Proposed Vehicle Model \label{subsec:Proposed-Modeling}}

\begin{figure*}[t]
\begin{centering}
\includegraphics[scale=0.085]{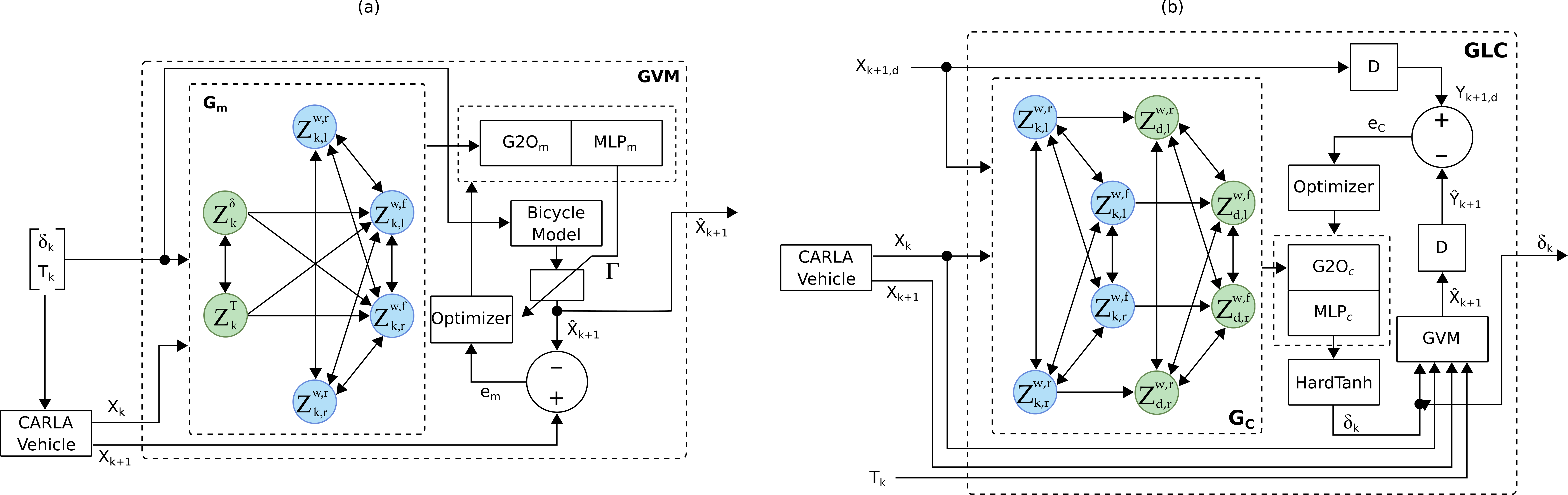}
\par\end{centering}
\begin{centering}
\vspace*{-1mm} 
\par\end{centering}
\caption{\textcolor{black}{(a) Architecture of Graph Vehicle Model to estimate
$\hat{X}_{k+1}$ online based on the error $e_{m}$ at each time step,
and, (b) Graph-based Lateral Controller framework for online lateral
control based on the error $e_{c}$ at each time step}\label{fig:Network-architecture}}
\end{figure*}

Since our proposed controller outputs the steering command and not
the actual states based on which the network needs to be optimized,
we need an intermediate differentiable and a high accuracy representation
of the actual vehicle. A commonly used model for ego cars, which is
also differentiable, is the dynamic bicycle model as shown in Figure
\ref{fig:Dynamic-vehicle-bicycle}, in which the front and rear wheel
pairs are each lumped into a single wheel since roll dynamics are
ignored. Assuming the center of mass is at the center of the wheelbase
(CW), the local frame of the vehicle is considered at CW. The dynamic
differential equations of the model are as follows:{\small{}
\begin{equation}
\dot{\hat{X}}_{b}=\left[\begin{array}{c}
\dot{x}\\
\dot{y}\\
\dot{\theta}\\
\dot{v}
\end{array}\right]=\left[\begin{array}{c}
v\mathrm{cos}\left(\theta+\beta\right)\\
v\mathrm{sin}\left(\theta+\beta\right)\\
\frac{v\mathrm{tan}\left(\delta\right)\mathrm{cos}\left(\beta\right)}{l_{\mathrm{wb}}}`\\
a_{\mathrm{net}}
\end{array}\right],\label{eq:kinematic_equations}
\end{equation}
}where, $x$ and $y$ are the global coordinates of the CW, $\theta$
is the heading, $v$ is the velocity of the vehicle, $\delta$ is
the steering angle, $\beta$ is the slip angle defined as{\small{}
\begin{equation}
\beta=\mathrm{arctan}\left(\frac{\mathrm{tan}\left(\delta\right)}{2}\right),
\end{equation}
}$l_{\mathrm{wb}}$ is the length of the wheelbase, and $a_{\mathrm{net}}$
is the net acceleration. Since, in CARLA, the vehicle takes throttle
($T$) as input instead of acceleration ($a$) and also experiences
drag force ($F_{\mathrm{drag}}$), the net acceleration can be defined
by{\small{}
\begin{equation}
a_{\mathrm{net}}=a-\frac{F_{\mathrm{drag}}}{m_{v}},\label{eq:acceleration_net}
\end{equation}
}with,{\small{}
\begin{equation}
F_{\mathrm{drag}}=\frac{1}{2}\zeta A_{f}\rho_{\mathrm{air}}v^{2},
\end{equation}
}and,{\small{}
\begin{equation}
a=6.5T^{2}+0.6T+0.08,\label{eq:acceleration}
\end{equation}
}where, $m_{v}$ and $\zeta$ are the mass and the drag coefficient
respectively obtained directly from CARLA, $A_{f}$ is the frontal
area that has been approximated based on the bounding box of the vehicle
provided in CARLA, and $\rho_{\mathrm{air}}$ is the is density. Equation
(\ref{eq:acceleration}) is obtained by a second-order polynomial
curve fitting of throttle versus acceleration data obtained through
simulation in CARLA.

The bicycle model described by equation (\ref{eq:kinematic_equations})
is limited by assumptions. Given a vehicle is a nonlinear dynamical
system running under uncertainties, the bicycle model cannot determine
the states accurately which is crucial for the controller performance.
Therefore, we propose the following Graph Vehicle Model (GVM) to estimate
the states:{\small{}
\begin{equation}
\dot{\chi}=\Gamma\odot\dot{\hat{X}}_{b},\label{eq:proposed_modeling}
\end{equation}
}where $\Gamma\in\mathbb{R}^{4\times1}$ allows dynamic model adjustment
and the $\odot$ operator represents element-wise multiplication.
\textcolor{black}{The mapping $\Gamma$ for vehicle modeling is learned
using $\mathrm{G2O}_{m}$ as follows:}{\small{}
\begin{equation}
\Gamma=\mathrm{MLP}_{m}\left(\left(\varphi_{m}\right)^{f}\right)\in\mathbb{R}^{4\times1},
\end{equation}
}\textcolor{black}{where, the subscript $m$ refers to modeling, $\varphi_{m}\in\mathbb{R}^{6\times M_{m}}$
(note that there are six nodes in the input graph $G_{m}$ which will
be explained later in this section), $\mathrm{MLP}$ refers to a deep
multilayer perceptron with input dimension $6M_{m}$, and, $\mathrm{G2O}$
and $\varphi$ have already been defined in section \ref{subsec:Graph-Neural-Network}.
Note that an MLP can be defined using the standard equation:}{\small{}
\begin{equation}
z^{(l)}=\sigma\left(W^{(l)}z^{(l-1)}+b^{(l)}\right),\textrm{ for }1\leq l\leq L,
\end{equation}
}\textcolor{black}{where, $z^{(0)}$ is the input, and $z^{(L)}$
is the output, $L$ is the number of layers including the input and
the output layers, and, $W^{(l)}$ and $b^{(l)}$ are its learnable
parameters in the $l^{\mathrm{th}}$ layer. The GVM architecture is
shown in Figure \ref{fig:Network-architecture}(a).}

The network is fed with a heterogeneous graph $G_{m}$, depicting
the four wheels and the actuators (steering and throttle) as nodes.
The features of the actuator nodes are their corresponding steering
and throttle values at $k^{\mathrm{th}}$ time $\left(Z_{k}^{\delta},Z_{k}^{T}\in\mathbb{R}\right)$.
The front wheel and the rear wheel nodes have the following features

{\small{}
\begin{equation}
Z_{k,l/r}^{w,f}=\left[\begin{array}{c}
s_{k,l/r}^{w,f}-s_{k}\\
d_{k,l/r}^{w,f}-d_{k}\\
v_{k}\\
\theta_{k}+\delta_{k-1}
\end{array}\right],\label{eq:wheel_features_front}
\end{equation}
}and,{\small{}
\begin{equation}
Z_{k,l/r}^{w,r}=\left[\begin{array}{c}
s_{k,l/r}^{w,r}-s_{k}\\
d_{k,l/r}^{w,r}-d_{k}\\
v_{k}\\
\theta_{k}
\end{array}\right],\label{eq:wheel_features_rear}
\end{equation}
}respectively, where $s$ and $d$ are the corresponding longitudinal
and lateral coordinates of $x$ and $y$ respectively in the Frenet
coordinate frame with respect to the reference path, the superscripts
$w$, $f$ and $r$ represent wheel, front and rear respectively,
and the subscripts $r$ and $l$ represent right and left respectively.
Thus, the first two elements of the feature vectors of the wheels
are in reference to CW. At the $k^{\mathrm{th}}$ step, the front
wheels are further angled by the steering angle at the $\left(k-1\right)^{\mathrm{th}}$
step. The wheels positions can be easily found using basic trigonometry
and their corresponding Frenet coordinates are computed. The edge
features between the wheels are the distances between them. Between
the actuator and the wheel nodes, the sampling period $t_{s}$ is
assigned as the edge feature. However, no edge feature is assigned
to edges between the actuator nodes themselves. The assigned edge
features are constant throughout time. Figure \ref{fig:Network-architecture}(a)
also shows the directions of the edges of $G_{m}$. It is not apparent
how the wheels would ``communicate'' with each other, and thus,
the edges between them are bidirectional. The same holds for the actuators.
Nonetheless, it is evident that the actuators drive the wheels --
both for steering and rotation. During implementation, the dynamics
of the model is discretized as follows:{\small{}
\begin{equation}
\hat{X}_{k+1}=X_{k}+\dot{\chi}_{k}t_{s},\label{eq:discretized_model}
\end{equation}
}where, $X$ is the ground truth of the states obtained from vehicle
sensors. The modeling error,{\small{}
\begin{equation}
e_{m}=X_{k+1}-\hat{X}_{k+1},
\end{equation}
}which is the difference between the estimated states and the ground
truth at the $\left(k+1\right)^{\mathrm{th}}$ time step, is backpropagated
into $\mathrm{G2O}_{m}$ and $\mathrm{MLP}_{m}$ to update their weights.

\subsection{Graph-based Lateral Controller \label{sec:Controller-Architecture}}

The Graph-based Lateral Controller (GLC) proposed in this section
primarily frames the ``desired'' transition from the current state
to the desired state into a heterogeneous graph $G_{c}$ as shown
in Figure \ref{fig:Network-architecture}(b). The features of the
wheels at $k^{\mathrm{th}}$ step are the same used in equations (\ref{eq:wheel_features_front})
and (\ref{eq:wheel_features_rear}). However, for the $\left(k+1\right)^{\mathrm{th}}$
step, velocity is removed from the feature array. The removal of velocity
from the desired nodes is intuitive since the task of the lateral
controller is not to track velocity profile. The unidirectional edges
from current to desired mimics the transition to the future, and thus,
the sampling period is assigned as their edge features. The bidirectional
edges between the wheels have their corresponding physical distances
as the edge feature.\textcolor{black}{{} As shown in Figure \ref{fig:Network-architecture}(b),
$G_{c}$ is fed into $\mathrm{G2O}_{c}$ to generate $\varphi_{c}$.
Finally, the steering command is produced as follows:}{\small{}
\begin{equation}
\delta_{k}=\mathrm{HardTanh}\left(\mathrm{MLP}_{c}\left(\left(\varphi_{c}^{\mathrm{desired}}\right)^{f}-\left(\varphi_{c}^{\mathrm{current}}\right)^{f}\right)\right),
\end{equation}
}\textcolor{black}{where, the subscript $c$ refers to controller,
$\varphi_{c}\in\mathbb{R}^{8\times M_{c}}$ (note that there are eight
nodes in $G_{c}$), $\mathrm{MLP}$ has an input dimension $4M_{c}$,
$\varphi_{c}^{\mathrm{desired}}$ and $\varphi_{c}^{\mathrm{current}}$
are the output node embedding of the desired and current nodes of
$G_{c}$ respectively, and,}{\small{}
\begin{equation}
\mathrm{HardTanh}\left(\omega\right)=\begin{cases}
1.0, & \textrm{if }\omega>1.0\\
-1.0, & \textrm{if }\omega<-1.0\\
2.0\omega, & \mathrm{otherwise}
\end{cases}.
\end{equation}
}Equation (\ref{eq:discretized_model}) is used to estimate the states
of the vehicle at $\left(k+1\right)$ using $\delta_{k}$ and compared
with the desired states. The error is then backpropagated into $\mathrm{G2O}_{c}$
and $\mathrm{MLP}_{c}$ to update their weights. To calculate the
error, only the position coordinates are used, i.e.,{\small{}
\begin{equation}
e_{c}=Y_{d}-\hat{Y}_{k+1}=Y_{d}-D\hat{X}_{k+1},
\end{equation}
}where, $Y_{d}$ is the desired position and $\hat{Y}$ is the estimated
position with $D=\left[\begin{array}{cc}
\boldsymbol{I}_{2\times2}, & \boldsymbol{0}_{2\times2}\end{array}\right]$ and $\boldsymbol{I}$ being the identity matrix. Algorithm \ref{alg:Procedure-Yd}
outlines the pseudo code to obtain $Y_{d}$. $C_{d}\left(R\right)$
is an array of the cumulative distance of the reference path $R$
given by positional coordinates, and the function ``Frenet Coordinate''
outputs the longitudinal Frenet coordinate with respect to the reference
path.

It is important to note that this work is centered on the design of
a lateral controller, and therefore, throttle is treated as a measured
disturbance in the vehicle dynamics. A similar approach was used in
the design of a lateral controller in \cite{Zhou2023}. 
\begin{figure}[H]
\begin{centering}
\includegraphics[scale=0.085]{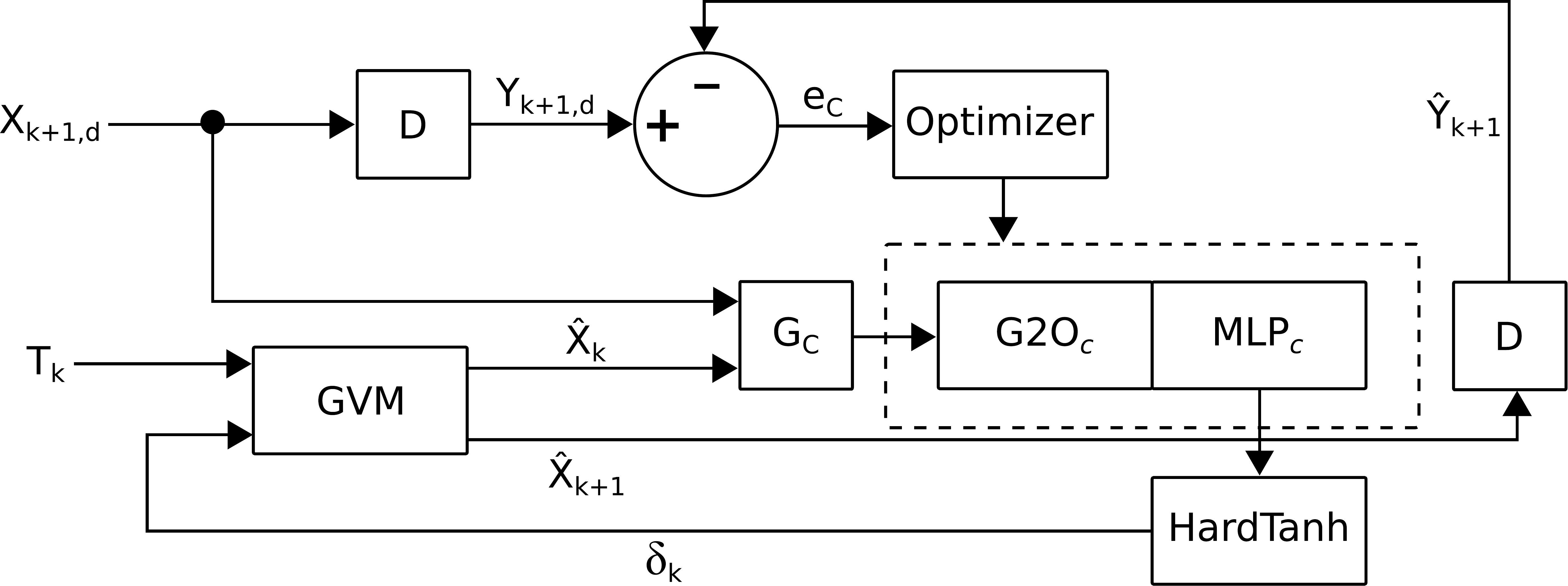}
\par\end{centering}
\begin{centering}
\vspace*{-1mm} 
\par\end{centering}
\caption{\textcolor{black}{Pre-training process of the controller \label{fig:Pre-training-controller}}}
\end{figure}
\begin{figure}[t]
\begin{centering}
\includegraphics[scale=0.3]{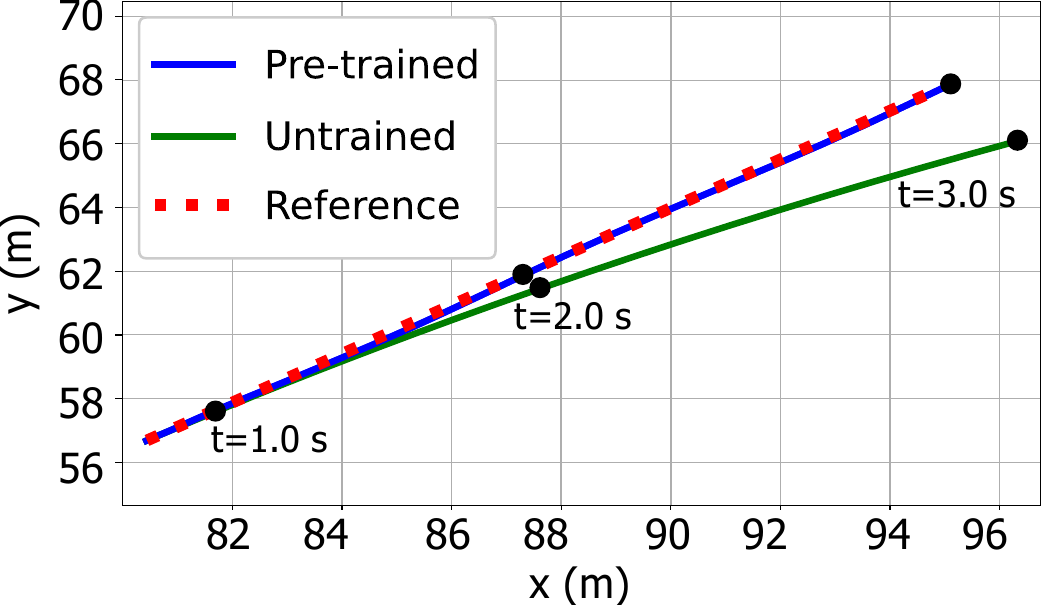}
\par\end{centering}
\begin{centering}
\vspace*{-1mm} 
\par\end{centering}
\caption{Performance comparison of the proposed controller in CARLA environment
at the first learning epoch with a pre-trained model and an untrained
model of the controller \label{fig:Pretrained_vs_untrained}}
\end{figure}

\section{Training Process\label{sec:Training-Process}}

Prior to deploying the online learning of the lateral controller,
GVM is pre-trained in the CARLA environment as shown in Figure \ref{fig:Network-architecture}.
By passing pre-determined throttle and steering values, the output
from the CARLA vehicle is utilized to train GVM on the fly. In practice,
a driver would just have to drive the car making as much diverse maneuvers
as possible and the network would learn in real time. 

Furthermore, we also pre-trained GLC on the already trained GVM before
the actual deployment in the CARLA simulator. The training process
is similar to Figure \ref{fig:Network-architecture}(b), except that
$X_{k}$ is replaced by $\hat{X}_{k}$, directly obtained from GVM,
to generate $G_{c}$ as shown in Figure \ref{fig:Pre-training-controller}.
The usefulness of this pre-training is evident in Figure \ref{fig:Pretrained_vs_untrained}
-- with a pre-trained model on the CARLA simulator, the vehicle starts
with impressive path tracking at the first learning epoch, unlike
when an untrained model is used. Again, in practicality, this would
mean that the car using a pre-trained GLC would be much safer at the
very beginning of its learning curve. It should be noted that during
the online learning, GVM can potentially be updated only when the
estimation error is above a predefined threshold $\varepsilon$. Similarly,
the online training of the GLC can be halted as long the performance
is satisfactory. However, in this work, the online learning is always
activated.

The loss function related to $e\in\mathbb{R}$ used to train both
the GVM and the GLC is SmoothL1Loss:{\small{}
\begin{equation}
\mathrm{smooth}_{\mathrm{L1}}\left(e\right):=\begin{cases}
0.5e^{2}, & \mathrm{if}\quad\left|e\right|<1\\
\left|e\right|-0.5, & \mathrm{otherwise}
\end{cases}.
\end{equation}
}
\begin{algorithm}[t]
\caption{Pseudo code to obtain $Y_{d}$ \label{alg:Procedure-Yd}}

\begin{lyxcode}
\textbf{\small{}Inputs}{\small{}:~$R$,~$C_{d}\left(R\right)$,~$X_{k}$}{\small\par}

\textbf{\small{}Do}{\small{}:}{\small\par}
\begin{lyxcode}
{\small{}1~$s_{k}=\mathrm{Frenet\:Coordinate}$$\left(R,X_{k}\right)$}{\small\par}

{\small{}2~$\mathrm{indices}=\mathrm{find}\left(C_{d}\left(R\right)>s_{k}\right)$~}{\small\par}

{\small{}3~$Y_{d}=R\left[\mathrm{indices[1]},:\right]$}{\small\par}
\end{lyxcode}
\end{lyxcode}
\end{algorithm}
\begin{table}[H]
\caption{MPC parameters and steering constraints used\label{tab:MPC-parameters}}

\centering{}{\footnotesize{}}%
\begin{tabular}{cc>{\centering}p{1in}}
\toprule 
\textbf{\footnotesize{}Parameter} & \textbf{\footnotesize{}Value} & \textbf{\footnotesize{}Description}\tabularnewline
\midrule 
{\footnotesize{}$p$} & {\footnotesize{}40} & {\footnotesize{}Number of time-steps in the prediction horizon}\tabularnewline
\midrule 
{\footnotesize{}$\delta_{\mathrm{min}}$} & {\footnotesize{}-1.0 rad} & {\footnotesize{}Minimum steering}\tabularnewline
\midrule 
{\footnotesize{}$\delta_{\mathrm{max}}$} & {\footnotesize{}1.0 rad} & {\footnotesize{}Maximum steering}\tabularnewline
\midrule 
{\footnotesize{}$\Delta\delta_{\mathrm{max}}$} & {\footnotesize{}0.5 rad/s} & {\footnotesize{}Maximum steering velocity}\tabularnewline
\midrule 
{\footnotesize{}$t_{s}$} & {\footnotesize{}0.1 s} & {\footnotesize{}Sampling period}\tabularnewline
\midrule 
{\footnotesize{}$R_{d}$} & {\footnotesize{}1.0} & {\footnotesize{}Weight on input}\tabularnewline
\midrule 
{\footnotesize{}$Q$} & {\footnotesize{}$\mathrm{diag}\left\{ \begin{array}{cc}
2.5, & 2.5\end{array}\right\} $} & {\footnotesize{}Weight on tracking error}\tabularnewline
\midrule 
{\footnotesize{}$Q_{f}$} & {\footnotesize{}$\mathrm{diag}\left\{ \begin{array}{cc}
3.5, & 3.5\end{array}\right\} $} & {\footnotesize{}Weight on terminal condition}\tabularnewline
\bottomrule
\end{tabular}{\footnotesize\par}
\end{table}

\section{Simulation Results \label{sec:Simulation-Results}}

\subsection{CARLA Environment Setup \label{sec:CARLA-Environment-Setup}}

\textcolor{black}{CARLA \cite{dosovitskiy2017carla} is an open-source
autonomous car simulator that can simulate driving scenarios and can
support multitude of ADS tasks \cite{li2024choose}. Its high-fidelity
environment allows to develop, test and validate ADS systems including
controllers. By default, CARLA employs the PhysX vehicle model by
NVIDIA and the model has been validated using ground-truth data from
an actual vehicle \cite{NVIDIA_PhysX}, making it a reasonable choice
for testing controllers. In recent works \cite{Zhou2023,yang2023real,li2023modified,pahk2024lane,khan2024simulation,yang2024e2e},
CARLA has been widely used for different tasks.}\textcolor{blue}{{}
}\textcolor{black}{T}he platform has the OPENDrive \cite{OpenDrive}
standalone mode that allows one to load an OpenDRIVE file and create
a temporal 3D mesh that defines the road in a minimalistic manner. 

In this work, CARLA version 0.9.14 has been used. At initialization,
the road is loaded using OpenDRIVE format along with the ego. The
center of the road along the path is set as the reference path. To
obtain the reference trajectory, the CARLA auto-pilot mode was activated
such that it follows the lane center at a maximum speed of $10\:\mathrm{m/s}$.
The reference trajectory contains position, heading and velocity at
each time-step, which is set at 0.1 second. CARLA APIs allow accessing
all the states (e.g., position, heading, velocity) directly at each
step. A longitudinal controller (PID) to follow the reference velocity,
and a lateral controller (the proposed controller) to minimize lateral
positional error, were implemented.

\textcolor{black}{The first autonomous racing leagues took place in
Abu Dhabi (circuit called Yas Marina, also used in this work) in April
2024 with seven participating nations \cite{a2rl}. At the trials
of the event during which no opponent was present on the track, the
self-driving cars randomly curled and even turned into the walls \cite{a2rl}
showing poor performances of the lateral component of the controllers.
Therefore, for training and testing purposes in this work, six different
Formula 1 race tracks from six different continents, as shown in Figure
\ref{fig:racing_tracks}, were used. These tracks have several sharp
and long turns, bends and straights making them practical choices
to test any control algorithm for autonomous cars.}

\subsection{Baselines}

\subsubsection{Model Predictive Controller}

A Linear Time-Varying Model Predictive Controller (MPC) is used as
the baseline lateral controller to compare our proposed controller
against. Since the prediction model (see equation (\ref{eq:kinematic_equations}))
is non-linear, it is linearized at different operating points, i.e.,
it is updated recursively as the operating conditions change. The
cost function to be minimized in the finite-horizon MPC problem is
defined using weighted 2-norms as follows:{\small{}
\begin{equation}
\begin{aligned}J= & \begin{aligned}\stackrel[k=1]{p-1}{\sum}\left\Vert Y_{t+k}-R_{t+k}\right\Vert _{Q}^{2} & +\end{aligned}
\begin{aligned}\left\Vert Y_{t+p}-R_{t+p}\right\Vert _{Q_{f}}^{2}\end{aligned}
\\
 & +\stackrel[k=1]{p-1}{\sum}\left\Vert \delta_{t+k}\right\Vert _{R_{d}}^{2}+\begin{aligned}\stackrel[k=1]{p-1}{\sum}\left\Vert \delta_{t+k}-\delta_{t+k-1}\right\Vert _{R_{d}}^{2}\end{aligned}
\end{aligned}
,
\end{equation}
}where, $p$ is the number of time-steps in the prediction horizon,
the first term represents the tracking error \textcolor{black}{of
the position and is penalized using the weight $Q$, the second term
penalizes the error of the terminal condition and is penalized using
the weight $Q_{f}$, the third term and the last term penalize high
steering inputs and steering velocity respectively, using the weight
$R_{d}$. The MPC is programmed to generate only the steering command
and a separate }PID is used as the longitudinal controller in the
baseline settings. The MPC parameters and constraints on the inputs
used are listed in Table \ref{tab:MPC-parameters}.

\subsubsection{\textcolor{black}{Stanley Lateral Controller}}

\textcolor{black}{The Stanley lateral controller \cite{thrun2006stanley}
uses position of the front axle to calculate the heading error ($\psi$)
as well as the cross-track error ($e_{\mathrm{ct}}$, the closest
distance between the front axle and the reference trajectory). The
steering angle (control law) is given as follows:}{\small{}
\begin{equation}
\delta=\psi+\mathrm{arctan}\frac{ke_{\mathrm{ct}}}{v},
\end{equation}
}\textcolor{black}{where, $k$ is a gain that determines the rate
of error convergence and $v$ is the velocity. The second term adjusts
the steering in nonlinear proportion to $e_{\mathrm{ct}}$, i.e.,
the steering response toward the trajectory is stronger if $e_{\mathrm{ct}}$
is larger. Detailed explanation on the Stanley lateral controller
can be found in \cite{thrun2006stanley}. In this work, $k=1.0$ has
been used.}

\subsection{Result Analysis}

The proposed controller GLC, the MPC and the Stanley controller are
all simulated in the CARLA environment as described earlier with the
goal to drive along the center lane. Maximum lateral error{\small{}
\begin{equation}
\mathrm{MLE}=\underset{i=1,\ldots,n}{\mathrm{max}}\left(\left|e_{i}\right|\right),
\end{equation}
}and the root mean square error{\small{}
\begin{equation}
\mathrm{RMSE}=\sqrt{\frac{1}{n}\stackrel[i=1]{n}{\sum}e_{i}^{2}},
\end{equation}
}
\begin{figure}[t]
\begin{centering}
\includegraphics[scale=0.25]{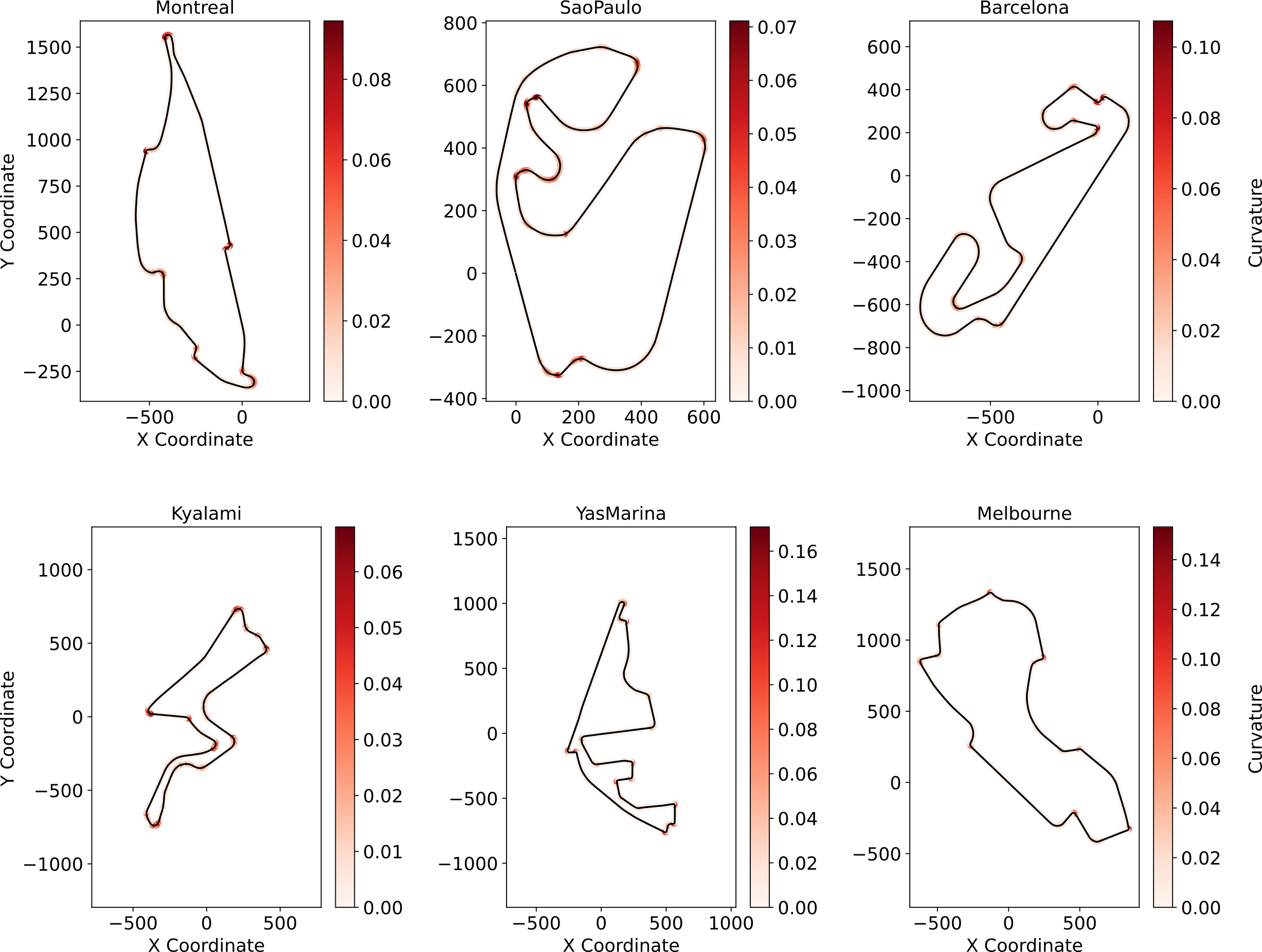}
\par\end{centering}
\caption{\textcolor{black}{Six race tracks from six different continents along
with curvature values along the race tracks\label{fig:racing_tracks}}}
\end{figure}
\begin{figure*}[t]
\begin{centering}
\includegraphics[scale=0.22]{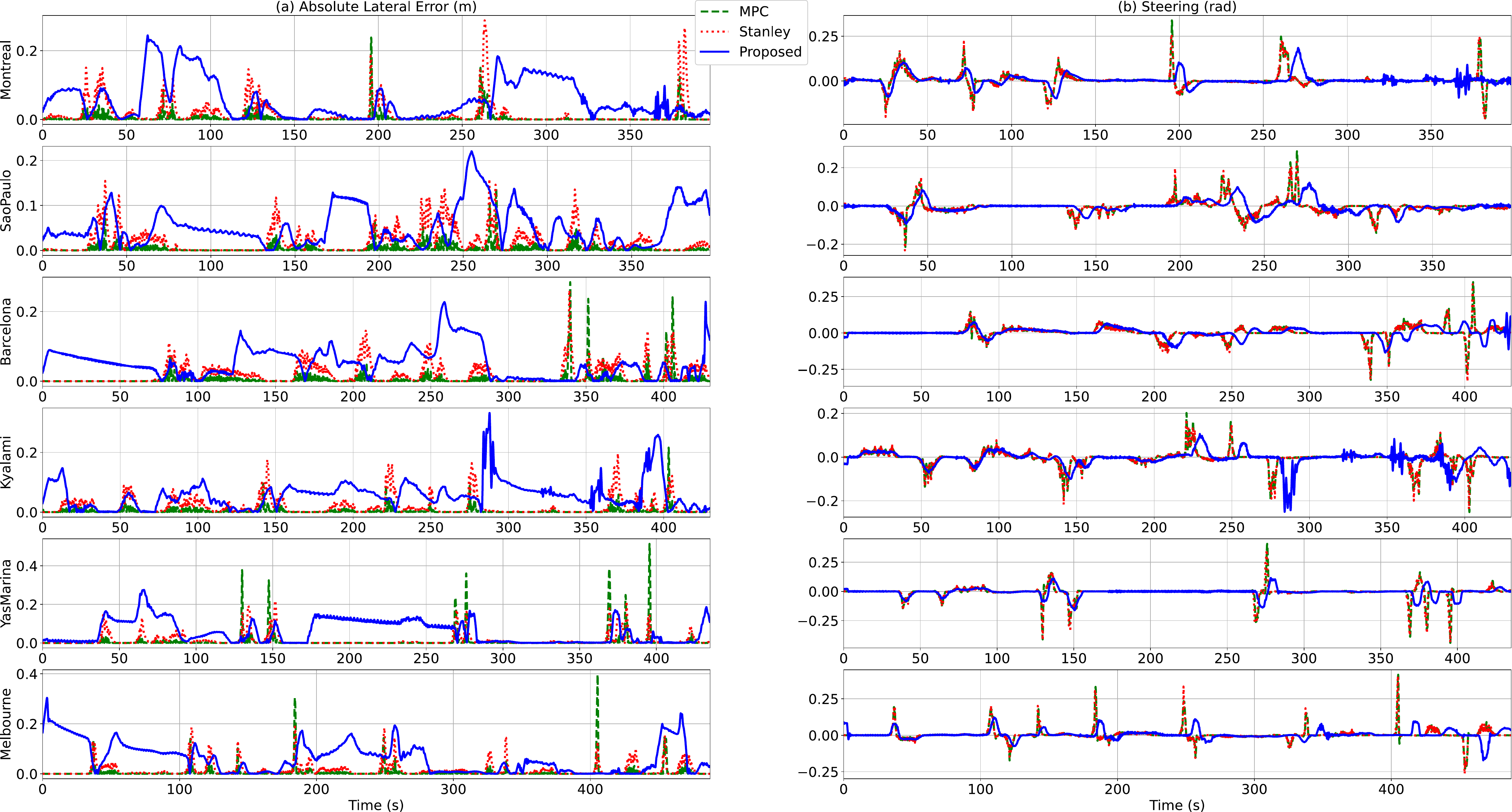}
\par\end{centering}
\begin{centering}
\vspace*{-1mm} 
\par\end{centering}
\caption{\textcolor{black}{Without perturbation -- (a) Absolute lateral error
experienced by the controllers, and (b) Steering commands generated
by the controllers \label{fig:Lateral-errorr-wo-perturb}}}
\end{figure*}
of the lateral position error are used to measure the performance
of each controller, where $e$ is the error signal for time $t_{1},\ldots,t_{n}$.
The tracks are segmented into two categories for better insight into
performance measurement: straight section when the curvature is \textcolor{black}{less
than $0.03\:\mathrm{m}^{-1}$, otherwise, turning section (see Figure
\ref{fig:racing_tracks}). Furthermore, with the pre-trained controller
discussed in section \ref{sec:Training-Process}, two different types
of tests -- with and without perturbation -- are carried out. For
the latter test, the controller is examined under two different perturbations:
(1) a constant perturbation of \textpm{} 2.5 degrees is added to the
steering command -- such a disturbance mimics the misalignment of
the wheels with the steering in the real world, and, (2) the friction
on the tires are reduced by 50\% from the default value, mimicking
a more slippery and a sudden change in the road surface. For simplicity,
we will refer to the former disturbance as D1 and the latter as D2.
Finally, tests are conducted to assess the generalization capability
of the proposed controller, employing three distinct car models -
Tesla Model 3, Toyota Prius, and Ford Mustang - obtained from the
CARLA simulator. However, in the context of comparing with the baselines,
only the Tesla Model 3 is employed for simplicity of presentation.}

\textcolor{black}{Firstly, the controllers are tested without perturbation.
Figure \ref{fig:Lateral-errorr-wo-perturb}(a) shows the performance
of our controller in contrast to the baselines and the summary of
the results are listed in Table \ref{tab:comparison-wo-perturb}.
On straight paths, MPC has outperformed both GLC and Stanley in terms
of both MLE and RMSE. At turns, GLC achieves better MLE in two of
the six tracks, while performance of MPC and Stanley are better in
one and three of six tracks, respectively. Overall, the RMSE values
show that the performance of the baselines are better than the proposed
controller without perturbation. However, this performance comes at
the cost of comfort. Although the steering commands generated by the
controllers, as shown in Figure \ref{fig:Lateral-errorr-wo-perturb}(b),
are within the steering limit of CARLA environment (i.e., $\textrm{1 radian}$),
Table \ref{tab:Steering_rate} shows that the maximum steering velocity
for all the six tracks generated by both MPC and Stanley is always
above GLC. Thus, our proposed controller provides a more comfortable
ride. }
\begin{table*}[!tp]
\caption{\textcolor{black}{Performance comparison for reference path tracking
between GLC, MPC and Stanley for the six tracks without perturbation.
The best result for each of the subcategories is in bold.}\label{tab:comparison-wo-perturb}}

\centering{}{\footnotesize{}}%
\begin{tabular}{cccccccccc}
\toprule 
\multirow{2}{*}{\textbf{\footnotesize{}MLE (m)}} & \multicolumn{3}{c}{\textbf{\footnotesize{}Straight}} & \multicolumn{3}{c}{\textbf{\footnotesize{}Turn}} & \multicolumn{3}{c}{\textbf{\footnotesize{}Overall}}\tabularnewline
\cmidrule{2-10} \cmidrule{3-10} \cmidrule{4-10} \cmidrule{5-10} \cmidrule{6-10} \cmidrule{7-10} \cmidrule{8-10} \cmidrule{9-10} \cmidrule{10-10} 
 & \textbf{\footnotesize{}GLC} & \textbf{\footnotesize{}MPC} & \textbf{\footnotesize{}Stanley} & \textbf{\footnotesize{}GLC} & \textbf{\footnotesize{}MPC} & \textbf{\footnotesize{}Stanley} & \textbf{\footnotesize{}GLC} & \textbf{\footnotesize{}MPC} & \textbf{\footnotesize{}Stanley}\tabularnewline
\midrule 
\textbf{\footnotesize{}Montreal} & {\footnotesize{}0.2444} & \textbf{\footnotesize{}0.0856} & {\footnotesize{}0.2589} & \textbf{\footnotesize{}0.1975} & {\footnotesize{}0.2382} & {\footnotesize{}0.2880} & {\footnotesize{}0.2444} & \textbf{\footnotesize{}0.2382} & {\footnotesize{}0.2880}\tabularnewline
\midrule 
\textbf{\footnotesize{}Sao Paulo} & {\footnotesize{}0.2206} & \textbf{\footnotesize{}0.0659} & {\footnotesize{}0.1547} & {\footnotesize{}0.2054} & \textbf{\footnotesize{}0.1375} & {\footnotesize{}0.1589} & {\footnotesize{}0.2206} & \textbf{\footnotesize{}0.1375} & {\footnotesize{}0.1589}\tabularnewline
\midrule 
\textbf{\footnotesize{}Barcelona} & {\footnotesize{}0.2281} & {\footnotesize{}0.1791} & \textbf{\footnotesize{}0.1743} & \textbf{\footnotesize{}0.1705} & {\footnotesize{}0.2845} & {\footnotesize{}0.2612} & \textbf{\footnotesize{}0.2281} & {\footnotesize{}0.2845} & {\footnotesize{}0.2612}\tabularnewline
\midrule 
\textbf{\footnotesize{}Kyalami} & {\footnotesize{}0.2696} & \textbf{\footnotesize{}0.1501} & {\footnotesize{}0.1816} & {\footnotesize{}0.3321} & {\footnotesize{}0.2161} & \textbf{\footnotesize{}0.1960} & {\footnotesize{}0.3321} & {\footnotesize{}0.2161} & \textbf{\footnotesize{}0.1960}\tabularnewline
\midrule 
\textbf{\footnotesize{}Yas Marina} & {\footnotesize{}0.2753} & {\footnotesize{}0.3139} & \textbf{\footnotesize{}0.2152} & {\footnotesize{}0.2747} & {\footnotesize{}0.5138} & \textbf{\footnotesize{}0.2190} & {\footnotesize{}0.2753} & {\footnotesize{}0.5138} & \textbf{\footnotesize{}0.2190}\tabularnewline
\midrule 
\textbf{\footnotesize{}Melbourne} & {\footnotesize{}0.3040} & {\footnotesize{}0.2704} & \textbf{\footnotesize{}0.1835} & {\footnotesize{}0.2610} & {\footnotesize{}0.3964} & \textbf{\footnotesize{}0.1890} & {\footnotesize{}0.3040} & {\footnotesize{}0.3964} & \textbf{\footnotesize{}0.1890}\tabularnewline
\midrule 
\multirow{2}{*}{\textbf{\footnotesize{}RMSE (m)}} & \multicolumn{3}{c}{\textbf{\footnotesize{}Straight}} & \multicolumn{3}{c}{\textbf{\footnotesize{}Turn}} & \multicolumn{3}{c}{\textbf{\footnotesize{}Overall}}\tabularnewline
\cmidrule{2-10} \cmidrule{3-10} \cmidrule{4-10} \cmidrule{5-10} \cmidrule{6-10} \cmidrule{7-10} \cmidrule{8-10} \cmidrule{9-10} \cmidrule{10-10} 
 & \textbf{\footnotesize{}GLC} & \textbf{\footnotesize{}MPC} & \textbf{\footnotesize{}Stanley} & \textbf{\footnotesize{}GLC} & \textbf{\footnotesize{}MPC} & \textbf{\footnotesize{}Stanley} & \textbf{\footnotesize{}GLC} & \textbf{\footnotesize{}MPC} & \textbf{\footnotesize{}Stanley}\tabularnewline
\midrule 
\textbf{\footnotesize{}Montreal} & {\footnotesize{}0.0813} & \textbf{\footnotesize{}0.0067} & {\footnotesize{}0.0310} & {\footnotesize{}0.0776} & \textbf{\footnotesize{}0.0728} & {\footnotesize{}0.1477} & {\footnotesize{}0.0809} & \textbf{\footnotesize{}0.0158} & {\footnotesize{}0.0416}\tabularnewline
\midrule 
\textbf{\footnotesize{}Sao Paulo} & {\footnotesize{}0.0681} & \textbf{\footnotesize{}0.0069} & {\footnotesize{}0.0310} & {\footnotesize{}0.0788} & \textbf{\footnotesize{}0.0554} & {\footnotesize{}0.0920} & {\footnotesize{}0.0688} & \textbf{\footnotesize{}0.0103} & {\footnotesize{}0.0334}\tabularnewline
\midrule 
\textbf{\footnotesize{}Barcelona} & {\footnotesize{}0.0675} & \textbf{\footnotesize{}0.0094} & {\footnotesize{}0.0314} & \textbf{\footnotesize{}0.0658} & {\footnotesize{}0.1253} & {\footnotesize{}0.1128} & {\footnotesize{}0.0674} & \textbf{\footnotesize{}0.0206} & {\footnotesize{}0.0351}\tabularnewline
\midrule 
\textbf{\footnotesize{}Kyalami} & {\footnotesize{}0.0664} & \textbf{\footnotesize{}0.0099} & {\footnotesize{}0.0318} & {\footnotesize{}0.1331} & \textbf{\footnotesize{}0.0593} & {\footnotesize{}0.0955} & {\footnotesize{}0.0744} & \textbf{\footnotesize{}0.0134} & {\footnotesize{}0.0344}\tabularnewline
\midrule 
\textbf{\footnotesize{}Yas Marina} & {\footnotesize{}0.0806} & \textbf{\footnotesize{}0.0102} & {\footnotesize{}0.0273} & {\footnotesize{}0.1090} & {\footnotesize{}0.1845} & \textbf{\footnotesize{}0.1042} & {\footnotesize{}0.0824} & {\footnotesize{}0.0405} & \textbf{\footnotesize{}0.0346}\tabularnewline
\midrule 
\textbf{\footnotesize{}Melbourne} & {\footnotesize{}0.0850} & \textbf{\footnotesize{}0.0087} & {\footnotesize{}0.0254} & {\footnotesize{}0.1216} & {\footnotesize{}0.1314} & \textbf{\footnotesize{}0.1054} & {\footnotesize{}0.0871} & \textbf{\footnotesize{}0.0246} & {\footnotesize{}0.0308}\tabularnewline
\bottomrule
\end{tabular}{\footnotesize\par}
\end{table*}
\begin{table}[t]
\caption{\textcolor{black}{Comparison of maximum steering velocity $\left(\dot{\delta}_{\mathrm{max}}\right)$
for reference path tracking between GLC, MPC and Stanley for the six
tracks. The best result for each of the subcategories is in bold.\label{tab:Steering_rate}}}

\centering{}{\footnotesize{}}%
\begin{tabular}{>{\raggedright}m{0.5in}>{\centering}p{0.15in}>{\centering}p{0.15in}>{\raggedright}m{0.25in}>{\centering}p{0.1in}>{\centering}p{0.1in}>{\centering}p{0.1in}>{\centering}p{0.1in}>{\centering}p{0.1in}>{\centering}p{0.1in}}
\toprule 
\multirow{3}{0.5in}{\textbf{\footnotesize{}$\dot{\delta}_{\mathrm{max}}$ (rad/s)}} & \multicolumn{3}{c}{\textbf{\footnotesize{}No Perturbation}} & \multicolumn{6}{c}{\textbf{\footnotesize{}With Perturbation}}\tabularnewline
\cmidrule{2-10} \cmidrule{3-10} \cmidrule{4-10} \cmidrule{5-10} \cmidrule{6-10} \cmidrule{7-10} \cmidrule{8-10} \cmidrule{9-10} \cmidrule{10-10} 
 & \multirow{2}{0.15in}{\textbf{\footnotesize{}GLC}} & \multirow{2}{0.15in}{\textbf{\footnotesize{}MPC}} & \multirow{2}{0.25in}{\textbf{\footnotesize{}Stanley}} & \multicolumn{2}{c}{\textbf{\footnotesize{}GLC}} & \multicolumn{2}{c}{\textbf{\footnotesize{}MPC}} & \multicolumn{2}{c}{\textbf{\footnotesize{}Stanley}}\tabularnewline
\cmidrule{5-10} \cmidrule{6-10} \cmidrule{7-10} \cmidrule{8-10} \cmidrule{9-10} \cmidrule{10-10} 
 &  &  &  & \textbf{\footnotesize{}D1} & \textbf{\footnotesize{}D2} & \textbf{\footnotesize{}D1} & \textbf{\footnotesize{}D2} & \textbf{\footnotesize{}D1} & \textbf{\footnotesize{}D2}\tabularnewline
\midrule 
\textbf{\footnotesize{}Montreal} & \textbf{\footnotesize{}0.25} & {\footnotesize{}0.61} & {\footnotesize{}0.95} & \textbf{\footnotesize{}0.17} & \textbf{\footnotesize{}0.13} & {\footnotesize{}1.01} & {\footnotesize{}1.15} & {\footnotesize{}1.41} & {\footnotesize{}0.70}\tabularnewline
\midrule 
\textbf{\footnotesize{}Sao Paulo} & \textbf{\footnotesize{}0.21} & {\footnotesize{}0.71} & {\footnotesize{}1.34} & \textbf{\footnotesize{}0.23} & \textbf{\footnotesize{}0.12} & {\footnotesize{}0.68} & {\footnotesize{}0.44} & {\footnotesize{}1.35} & {\footnotesize{}0.77}\tabularnewline
\midrule 
\textbf{\footnotesize{}Barcelona} & \textbf{\footnotesize{}0.33} & {\footnotesize{}0.62} & {\footnotesize{}0.70} & \textbf{\footnotesize{}0.18} & \textbf{\footnotesize{}0.12} & {\footnotesize{}1.34} & {\footnotesize{}0.64} & {\footnotesize{}1.34} & {\footnotesize{}1.20}\tabularnewline
\midrule 
\textbf{\footnotesize{}Kyalami} & \textbf{\footnotesize{}0.26} & {\footnotesize{}0.61} & {\footnotesize{}0.80} & \textbf{\footnotesize{}0.28} & \textbf{\footnotesize{}0.14} & {\footnotesize{}0.73} & {\footnotesize{}0.50} & {\footnotesize{}1.03} & {\footnotesize{}0.71}\tabularnewline
\midrule 
\textbf{\footnotesize{}Yas Marina} & \textbf{\footnotesize{}0.12} & {\footnotesize{}2.20} & {\footnotesize{}1.04} & \textbf{\footnotesize{}0.26} & \textbf{\footnotesize{}0.25} & {\footnotesize{}1.66} & {\footnotesize{}0.84} & {\footnotesize{}1.04} & {\footnotesize{}1.64}\tabularnewline
\midrule 
\textbf{\footnotesize{}Melbourne} & \textbf{\footnotesize{}0.22} & {\footnotesize{}0.79} & {\footnotesize{}1.29} & \textbf{\footnotesize{}0.22} & \textbf{\footnotesize{}0.16} & {\footnotesize{}0.91} & {\footnotesize{}0.82} & {\footnotesize{}1.51} & {\footnotesize{}0.58}\tabularnewline
\bottomrule
\end{tabular}{\footnotesize\par}
\end{table}

\textcolor{black}{The performance of the proposed GVM plays a crucial
role in the performance of GLC. Table \ref{tab:rmse_position_estimation}
shows the accuracy of our model in estimating the position of the
ego. The RMSE values suggest high accuracy of our proposed vehicle
model.}
\begin{table}[t]
\caption{\textcolor{black}{RMSE for position estimation by GVM for the six
tracks\label{tab:rmse_position_estimation}}}

\centering{}{\footnotesize{}}%
\begin{tabular}{cc}
\toprule 
 & \multicolumn{1}{c}{\textbf{\footnotesize{}RMSE (m)}}\tabularnewline
\midrule 
\textbf{\footnotesize{}Montreal} & {\footnotesize{}0.021}\tabularnewline
\midrule 
\textbf{\footnotesize{}Sao Paulo} & {\footnotesize{}0.020}\tabularnewline
\midrule 
\textbf{\footnotesize{}Barcelona} & {\footnotesize{}0.018}\tabularnewline
\midrule 
\textbf{\footnotesize{}Kyalami} & {\footnotesize{}0.018}\tabularnewline
\midrule 
\textbf{\footnotesize{}Yas Marina} & {\footnotesize{}0.026}\tabularnewline
\midrule 
\textbf{\footnotesize{}Melbourne} & {\footnotesize{}0.019}\tabularnewline
\bottomrule
\end{tabular}{\footnotesize\par}
\end{table}

\begin{figure*}[t]
\begin{centering}
\includegraphics[scale=0.22]{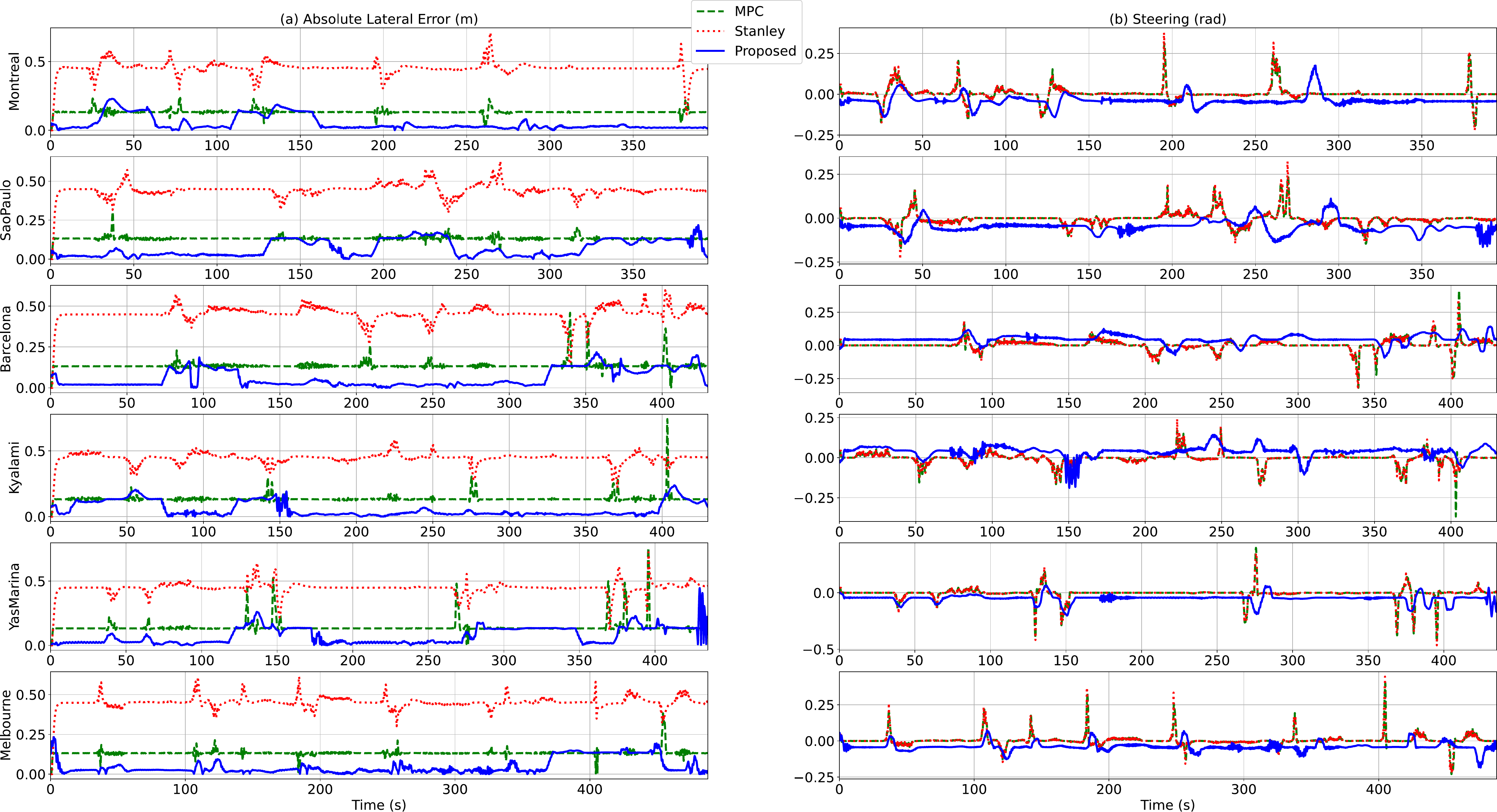}
\par\end{centering}
\begin{centering}
\vspace*{-1mm} 
\par\end{centering}
\caption{\textcolor{black}{With perturbation D1 -- (a) Absolute lateral error
experienced by the controllers, and, (b) Steering commands generated
by the controllers \label{fig:Lateral-errorr-with-D1}}}
\end{figure*}

\begin{figure*}[t]
\begin{centering}
\includegraphics[scale=0.22]{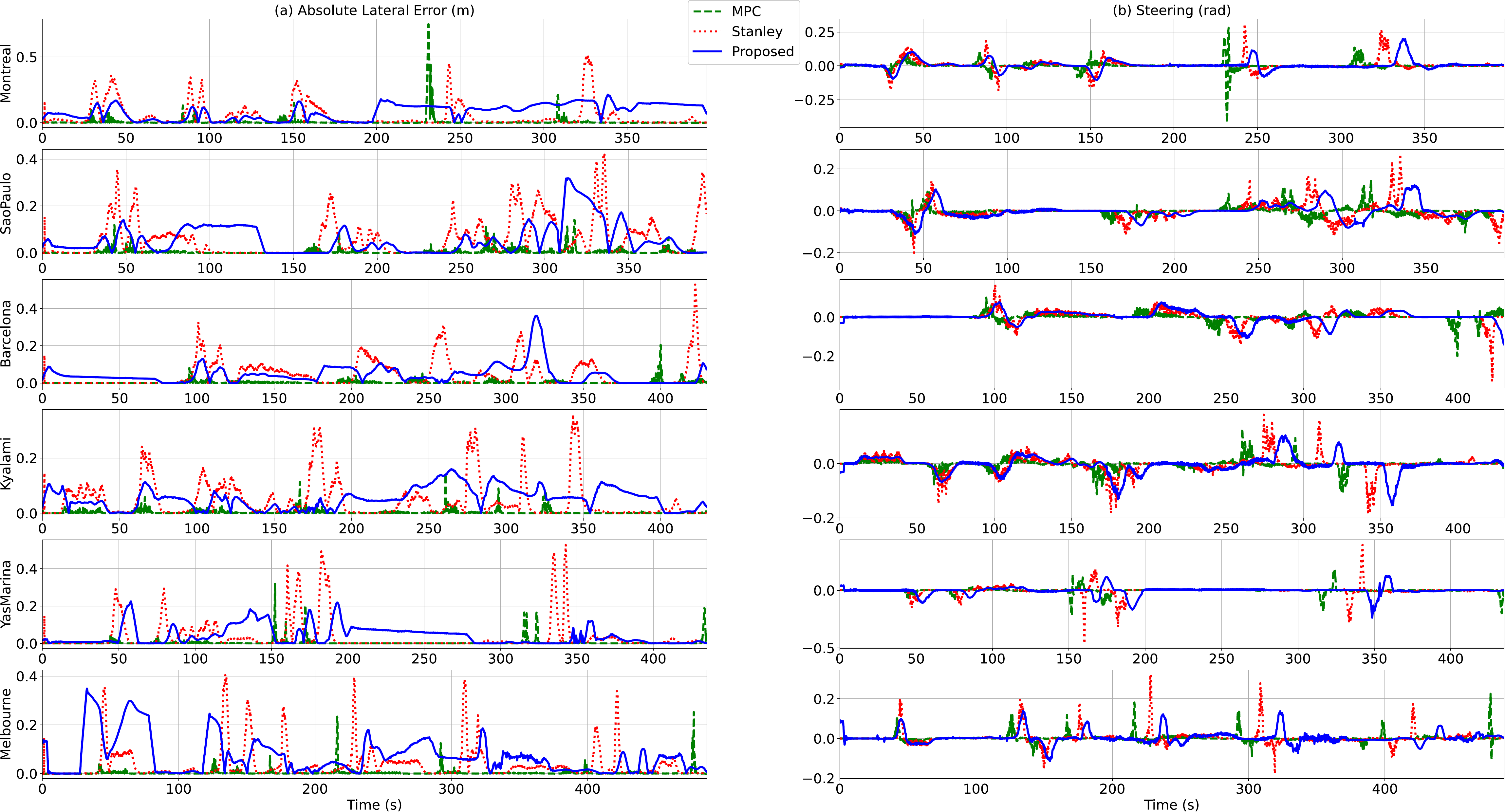}
\par\end{centering}
\begin{centering}
\vspace*{-1mm} 
\par\end{centering}
\caption{\textcolor{black}{With perturbation D2 -- (a) Absolute lateral error
experienced by the controllers, and, (b) Steering commands generated
by the controllers \label{fig:Lateral-errorr-with-D2}}}
\end{figure*}

\textcolor{black}{Next, the controllers are tested with perturbations
D1 and D2. It can be seen from Figure \ref{fig:Lateral-errorr-with-D1}(a)
that the proposed controller GLC learned to attenuate D1 while MPC
and Stanley failed to do so -- a lateral error offset can be observed
for both. In terms of both MLE and RMSE as can be seen in Table \ref{tab:comparison-perturb},
the performance of GLC is superior to that of MPC and Stanley when
perturbed with D1. With D2, MPC generally performs better than GLC
in majority of the tracks with respect to both RMSE and MLE as can
be seen in Figure \ref{fig:Lateral-errorr-with-D2}(a) and Table \ref{tab:comparison-perturb}.
However, MPC shows wobbling behavior at some of the turns, e.g., between
$229\textrm{s}$ and $334\textrm{s}$ in Montreal as can be seen in
Figure \ref{fig:Lateral-errorr-with-D2}(a). Such wobbling behavior
is undesirable and was absent in our proposed controller. Stanley,
on the other hand, had the least performance with D2 for all the six
tracks. Note that our controller maintained a more comfortable ride
even under perturbations (see Figure \ref{fig:Lateral-errorr-with-D1}(b),
Figure \ref{fig:Lateral-errorr-with-D2}(b) and Table \ref{tab:Steering_rate}).
Furthermore, in general, GLC had lower steering velocity under D2
in contrast to D1, showcasing its adaptability to learn to steer at
a slower rate on slippery roads and this behavior was also visible
in the other two controllers.}
\begin{table*}[t]
\caption{\textcolor{black}{Performance comparison for reference path tracking
between GLC, MPC and Stanley for the six tracks with perturbation.
The best result for each of the subcategories is in bold.\label{tab:comparison-perturb}}}

\centering{}{\footnotesize{}}%
\begin{tabular}{ccccccccccccc}
\toprule 
 & \multicolumn{6}{c}{\textbf{\footnotesize{}D1}} & \multicolumn{6}{c}{\textbf{\footnotesize{}D2}}\tabularnewline
\midrule 
\multirow{2}{*}{\textbf{\footnotesize{}MLE (m)}} & \multicolumn{3}{c}{\textbf{\footnotesize{}Turn}} & \multicolumn{3}{c}{\textbf{\footnotesize{}Overall}} & \multicolumn{3}{c}{\textbf{\footnotesize{}Turn}} & \multicolumn{3}{c}{\textbf{\footnotesize{}Overall}}\tabularnewline
\cmidrule{2-13} \cmidrule{3-13} \cmidrule{4-13} \cmidrule{5-13} \cmidrule{6-13} \cmidrule{7-13} \cmidrule{8-13} \cmidrule{9-13} \cmidrule{10-13} \cmidrule{11-13} \cmidrule{12-13} \cmidrule{13-13} 
 & \textbf{\footnotesize{}GLC} & \textbf{\footnotesize{}MPC} & \textbf{\footnotesize{}Stanley} & \textbf{\footnotesize{}GLC} & \textbf{\footnotesize{}MPC} & \textbf{\footnotesize{}Stanley} & \textbf{\footnotesize{}GLC} & \textbf{\footnotesize{}MPC} & \textbf{\footnotesize{}Stanley} & \textbf{\footnotesize{}GLC} & \textbf{\footnotesize{}MPC} & \textbf{\footnotesize{}Stanley}\tabularnewline
\midrule 
\textbf{\footnotesize{}Montreal} & \textbf{\footnotesize{}0.2286} & {\footnotesize{}0.2315} & {\footnotesize{}0.7093} & \textbf{\footnotesize{}0.2287} & {\footnotesize{}0.2354} & {\footnotesize{}0.7093} & \textbf{\footnotesize{}0.2124} & {\footnotesize{}0.7497} & {\footnotesize{}0.5157} & \textbf{\footnotesize{}0.2137} & {\footnotesize{}0.7497} & {\footnotesize{}0.5157}\tabularnewline
\midrule 
\textbf{\footnotesize{}Sao Paulo} & \textbf{\footnotesize{}0.2185} & {\footnotesize{}0.3031} & {\footnotesize{}0.6267} & \textbf{\footnotesize{}0.2185} & {\footnotesize{}0.3031} & {\footnotesize{}0.6267} & {\footnotesize{}0.2438} & \textbf{\footnotesize{}0.1413} & {\footnotesize{}0.4206} & {\footnotesize{}0.3190} & \textbf{\footnotesize{}0.1413} & {\footnotesize{}0.4206}\tabularnewline
\midrule 
\textbf{\footnotesize{}Barcelona} & \textbf{\footnotesize{}0.2183} & {\footnotesize{}0.4597} & {\footnotesize{}0.5973} & \textbf{\footnotesize{}0.2183} & {\footnotesize{}0.4597} & {\footnotesize{}0.5973} & {\footnotesize{}0.3365} & \textbf{\footnotesize{}0.2054} & {\footnotesize{}0.5286} & {\footnotesize{}0.3615} & \textbf{\footnotesize{}0.2054} & {\footnotesize{}0.5286}\tabularnewline
\midrule 
\textbf{\footnotesize{}Kyalami} & \textbf{\footnotesize{}0.2356} & {\footnotesize{}0.7426} & {\footnotesize{}0.5895} & \textbf{\footnotesize{}0.2385} & {\footnotesize{}0.7426} & {\footnotesize{}0.5895} & {\footnotesize{}0.1578} & \textbf{\footnotesize{}0.1410} & {\footnotesize{}0.3569} & {\footnotesize{}0.1601} & \textbf{\footnotesize{}0.1410} & {\footnotesize{}0.3569}\tabularnewline
\midrule 
\textbf{\footnotesize{}Yas Marina} & \textbf{\footnotesize{}0.4074} & {\footnotesize{}0.7589} & {\footnotesize{}0.7248} & \textbf{\footnotesize{}0.4460} & {\footnotesize{}0.7589} & {\footnotesize{}0.7248} & \textbf{\footnotesize{}0.2221} & {\footnotesize{}0.3193} & {\footnotesize{}0.5275} & \textbf{\footnotesize{}0.2249} & {\footnotesize{}0.3193} & {\footnotesize{}0.5275}\tabularnewline
\midrule 
\textbf{\footnotesize{}Melbourne} & \textbf{\footnotesize{}0.2333} & {\footnotesize{}0.3830} & {\footnotesize{}0.6133} & \textbf{\footnotesize{}0.2333} & {\footnotesize{}0.3830} & {\footnotesize{}0.6133} & \textbf{\footnotesize{}0.2179} & {\footnotesize{}0.2527} & {\footnotesize{}0.3970} & {\footnotesize{}0.3492} & \textbf{\footnotesize{}0.2527} & {\footnotesize{}0.4045}\tabularnewline
\midrule 
\multirow{2}{*}{\textbf{\footnotesize{}RMSE (m)}} & \multicolumn{3}{c}{\textbf{\footnotesize{}Turn}} & \multicolumn{3}{c}{\textbf{\footnotesize{}Overall}} & \multicolumn{3}{c}{\textbf{\footnotesize{}Turn}} & \multicolumn{3}{c}{\textbf{\footnotesize{}Overall}}\tabularnewline
\cmidrule{2-13} \cmidrule{3-13} \cmidrule{4-13} \cmidrule{5-13} \cmidrule{6-13} \cmidrule{7-13} \cmidrule{8-13} \cmidrule{9-13} \cmidrule{10-13} \cmidrule{11-13} \cmidrule{12-13} \cmidrule{13-13} 
 & \textbf{\footnotesize{}GLC} & \textbf{\footnotesize{}MPC} & \textbf{\footnotesize{}Stanley} & \textbf{\footnotesize{}GLC} & \textbf{\footnotesize{}MPC} & \textbf{\footnotesize{}Stanley} & \textbf{\footnotesize{}GLC} & \textbf{\footnotesize{}MPC} & \textbf{\footnotesize{}Stanley} & \textbf{\footnotesize{}GLC} & \textbf{\footnotesize{}MPC} & \textbf{\footnotesize{}Stanley}\tabularnewline
\midrule 
\textbf{\footnotesize{}Montreal} & \textbf{\footnotesize{}0.0575} & {\footnotesize{}0.1482} & {\footnotesize{}0.4884} & \textbf{\footnotesize{}0.0719} & {\footnotesize{}0.1342} & {\footnotesize{}0.4530} & \textbf{\footnotesize{}0.1177} & {\footnotesize{}0.2110} & {\footnotesize{}0.3476} & {\footnotesize{}0.1013} & \textbf{\footnotesize{}0.0442} & {\footnotesize{}0.1023}\tabularnewline
\midrule 
\textbf{\footnotesize{}Sao Paulo} & \textbf{\footnotesize{}0.0700} & {\footnotesize{}0.1529} & {\footnotesize{}0.5009} & \textbf{\footnotesize{}0.0853} & {\footnotesize{}0.1337} & {\footnotesize{}0.4446} & {\footnotesize{}0.0924} & \textbf{\footnotesize{}0.0706} & {\footnotesize{}0.2849} & {\footnotesize{}0.0867} & \textbf{\footnotesize{}0.0130} & {\footnotesize{}0.0989}\tabularnewline
\midrule 
\textbf{\footnotesize{}Barcelona} & \textbf{\footnotesize{}0.0929} & {\footnotesize{}0.2478} & {\footnotesize{}0.4511} & \textbf{\footnotesize{}0.0780} & {\footnotesize{}0.1386} & {\footnotesize{}0.4559} & {\footnotesize{}0.0944} & \textbf{\footnotesize{}0.0786} & {\footnotesize{}0.3412} & {\footnotesize{}0.0674} & \textbf{\footnotesize{}0.0109} & {\footnotesize{}0.0832}\tabularnewline
\midrule 
\textbf{\footnotesize{}Kyalami} & \textbf{\footnotesize{}0.0707} & {\footnotesize{}0.2668} & {\footnotesize{}0.3958} & \textbf{\footnotesize{}0.0784} & {\footnotesize{}0.1388} & {\footnotesize{}0.4463} & {\footnotesize{}0.0813} & \textbf{\footnotesize{}0.0542} & {\footnotesize{}0.2631} & {\footnotesize{}0.0652} & \textbf{\footnotesize{}0.0093} & {\footnotesize{}0.0823}\tabularnewline
\midrule 
\textbf{\footnotesize{}Yas Marina} & \textbf{\footnotesize{}0.1444} & {\footnotesize{}0.3282} & {\footnotesize{}0.4562} & \textbf{\footnotesize{}0.0996} & {\footnotesize{}0.1487} & {\footnotesize{}0.4485} & \textbf{\footnotesize{}0.0928} & {\footnotesize{}0.1021} & {\footnotesize{}0.3493} & {\footnotesize{}0.0629} & \textbf{\footnotesize{}0.0205} & {\footnotesize{}0.0895}\tabularnewline
\midrule 
\textbf{\footnotesize{}Melbourne} & \textbf{\footnotesize{}0.0790} & {\footnotesize{}0.1747} & {\footnotesize{}0.4958} & \textbf{\footnotesize{}0.0674} & {\footnotesize{}0.1341} & {\footnotesize{}0.4522} & {\footnotesize{}0.0908} & \textbf{\footnotesize{}0.0841} & {\footnotesize{}0.2961} & {\footnotesize{}0.1031} & \textbf{\footnotesize{}0.0144} & {\footnotesize{}0.0756}\tabularnewline
\bottomrule
\end{tabular}{\footnotesize\par}
\end{table*}

\textcolor{black}{The performance evaluation of GLC across the six
racetracks was conducted using three car models and the results are
summarized in Table \ref{tab:performances-cars}. Among these, the
Toyota Prius consistently demonstrated the best performance across
all tracks. The Ford Mustang, however, performed the worst in terms
of RMSE and MLE on most tracks, except for two instances. On the Kyalami
track, the Tesla Model 3 recorded the highest MLE of $0.3321\textrm{m}$,
and on the Yas Marina track, it also showed the worst RMSE of $0.0823\textrm{m}$.
In comparison, the Ford Mustang averaged an MLE of $0.3485\textrm{m}$
and an RMSE of $0.1126\textrm{m}$ across the remaining tracks, indicating
that the adaptability of GLC was less effective for the dynamics of
the Ford Mustang, while it achieved more reliable performance with
the Toyota Prius. Overall, the generalization capability of GLC on
different car models is satisfactory from the results.}
\begin{table*}[t]
\caption{\textcolor{black}{Overall performance of GLC on different car for
the six racetracks. The worst result for each of the racetracks is
in bold. \label{tab:performances-cars}}}

\centering{}%
\begin{tabular}{c>{\centering}p{0.5in}>{\centering}p{0.5in}>{\centering}p{0.5in}>{\centering}p{0.5in}>{\centering}p{0.5in}>{\centering}p{0.5in}}
\toprule 
\multirow{2}{*}{} & \multicolumn{3}{c}{\textbf{\footnotesize{}MLE (m)}} & \multicolumn{3}{c}{\textbf{\footnotesize{}RMSE (m)}}\tabularnewline
\cmidrule{2-7} \cmidrule{3-7} \cmidrule{4-7} \cmidrule{5-7} \cmidrule{6-7} \cmidrule{7-7} 
 & \textbf{\textcolor{black}{\footnotesize{}Tesla Model 3}} & \textbf{\textcolor{black}{\footnotesize{}Toyota Prius}} & \textbf{\textcolor{black}{\footnotesize{}Ford Mustang}} & \textbf{\textcolor{black}{\footnotesize{}Tesla Model 3}} & \textbf{\textcolor{black}{\footnotesize{}Toyota Prius}} & \textbf{\textcolor{black}{\footnotesize{}Ford Mustang}}\tabularnewline
\midrule 
\textbf{\footnotesize{}Montreal} & {\footnotesize{}0.2444} & {\footnotesize{}0.2204} & \textbf{\footnotesize{}0.4872} & {\footnotesize{}0.0809} & {\footnotesize{}0.1113} & \textbf{\footnotesize{}0.1460}\tabularnewline
\midrule 
\textbf{\footnotesize{}Sao Paulo} & {\footnotesize{}0.2206} & {\footnotesize{}0.1580} & \textbf{\footnotesize{}0.3747} & {\footnotesize{}0.0688} & {\footnotesize{}0.0673} & \textbf{\footnotesize{}0.1364}\tabularnewline
\midrule 
\textbf{\footnotesize{}Barcelona} & {\footnotesize{}0.2281} & {\footnotesize{}0.1280} & \textbf{\footnotesize{}0.2986} & {\footnotesize{}0.0674} & {\footnotesize{}0.0458} & \textbf{\footnotesize{}0.0777}\tabularnewline
\midrule 
\textbf{\footnotesize{}Kyalami} & \textbf{\footnotesize{}0.3321} & {\footnotesize{}0.1961} & {\footnotesize{}0.2431} & {\footnotesize{}0.0744} & {\footnotesize{}0.0726} & \textbf{\footnotesize{}0.0883}\tabularnewline
\midrule 
\textbf{\footnotesize{}Yas Marina} & {\footnotesize{}0.2753} & {\footnotesize{}0.2316} & \textbf{\footnotesize{}0.2874} & \textbf{\footnotesize{}0.0823} & {\footnotesize{}0.0726} & {\footnotesize{}0.0728}\tabularnewline
\midrule 
\textbf{\footnotesize{}Melbourne} & {\footnotesize{}0.3040} & {\footnotesize{}0.3361} & \textbf{\footnotesize{}0.3997} & {\footnotesize{}0.0849} & {\footnotesize{}0.1015 } & \textbf{\footnotesize{}0.1543}\tabularnewline
\bottomrule
\end{tabular}
\end{table*}

\textcolor{black}{All simulations were conducted on a PC equipped
with an Intel Core i9-10900F CPU processor, 32 GB of RAM, and an NVIDIA
GeForce RTX 3070 GPU, running on Windows 11. For the experiments conducted,
the average inference time and the average training time of GLC are
$15\textrm{ms}$ and $29\textrm{ms}$ respectively, which are within
the acceptable range in the area of autonomous vehicles \cite{biglari2023review}.}

\section{Conclusion \label{sec:Conclusion}}

This work introduces an innovative approach of learning a vehicle
model followed by a novel online lateral controller for autonomous
driving. We use heterogeneous graphs to represent both the components
and then feed them through their corresponding G2O. The vehicle learning
model GVM utilizes existing knowledge of a vehicle, whereas a neural
network would have to learn from scratch. Conversely, unlike a physics-based
model, GVM can capture unidentified dynamics. Combining these strengths,
GVM efficiently learned the vehicle dynamics online. \textcolor{black}{The
lateral controller GLC also learns online to effectively maintain
the ego on its reference path; MPC and Stanley controller, on the
other hand, cannot learn online. Without perturbation, MPC and Stanley
does better tracking than GLC. However, the proposed controller gives
a more comfortable ride. When the ego is perturbed with D1, GLC outperforms
both MPC and Stanley by mitigating the perturbation while maintaining
satisfactory tracking performance. Nonetheless, when the tires are
perturbed by D2, MPC generally performed better than GLC, despite
occasionally exhibiting undesired wobbling behavior not observed in
GLC. These experimental results provide evidence of the robustness
of GLC in the presence of both the perturbations, while MPC and Stanley
were inadequate to attenuate D1. Furthermore, we also showed the generalization
capability of GLC with different car models while achieving satisfactory
performance. }

The efficacy and practicality of our proposed techniques have been
validated through comprehensive simulations in CARLA. In this work,
we tested our lateral controller for a low velocity and a PID controller
was used as a longitudinal controller for tracking that velocity.
However, at higher velocities, the coupling between the two controllers
become stronger, and thus, a unified controller is more suitable.
For future work, we want to investigate a graph-based controller that
can generate both the steering and the throttle commands.

\bibliographystyle{unsrt}
\bibliography{VehicleControl_Reference}

\section*{Biography }

\vspace{-2pt}

\begin{IEEEbiography}[{\includegraphics[width=1in,height=1.25in]{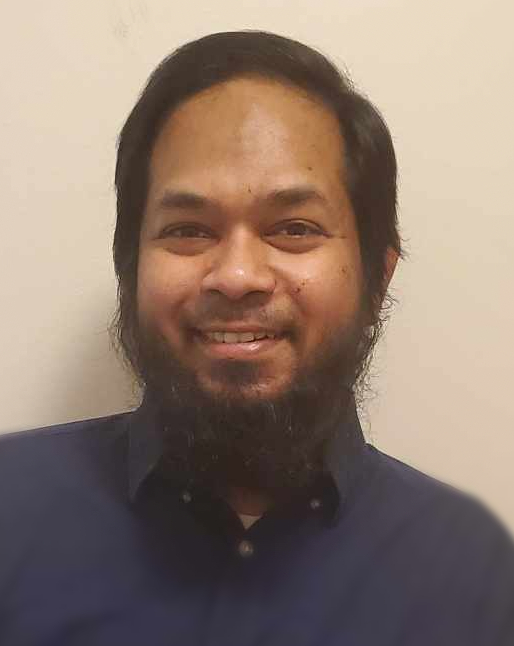}}]{Jilan Samiuddin}
  received his B.Sc. degree in Electrical and Electronic Engineering
from American International University - Bangladesh and M.Sc. degree
in Electrical Engineering from the University of Calgary in 2012 and
2016, respectively. He is currently a Ph.D. candidate in the Department
of Electrical and Computer Engineering at McGill University. His research
is primarily focused on machine learning algorithms with the motivation
of developing novel strategies for autonomous driving.
\end{IEEEbiography}

\vspace{-2pt}

\begin{IEEEbiography}[{\includegraphics[width=1in,height=1.25in]{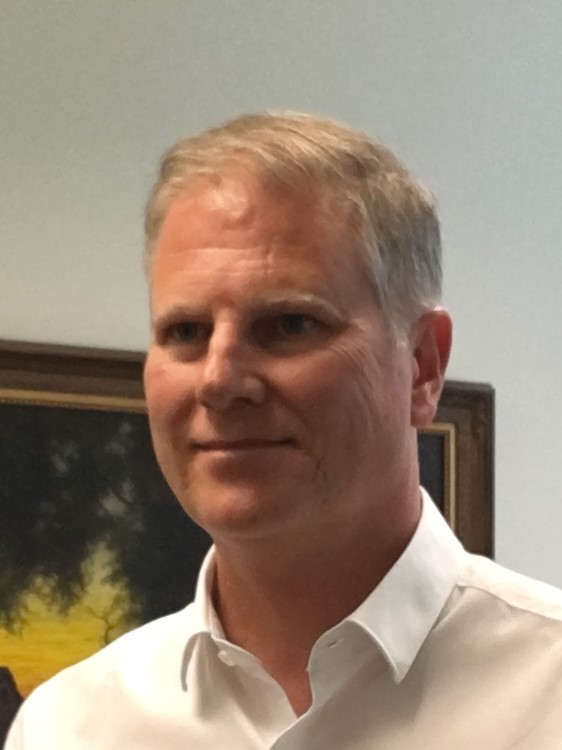}}]{Benoit Boulet}
 , P.Eng., Ph.D., is Professor of Electrical and Computer Engineering
at McGill University which he joined in 1998, and Director of the
McGill Engine. He is also McGill\textquoteright s Associate Vice-President
(Innovation and Partnerships). Prof. Boulet obtained an M.Eng. degree
from McGill in 1992 and a Ph.D. degree from the University of Toronto
in 1996, both in electrical engineering. He is a former Director and
current member of the Centre for Intelligent Machines where he heads
the Intelligent Automation Laboratory. His research interests include
the design and data-driven control of autonomous electric vehicles
and renewable energy systems and applications of machine learning.
\end{IEEEbiography}

\vspace{-2pt}

\begin{IEEEbiography}[{\includegraphics[width=1in,height=1.25in]{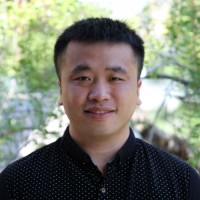}}]{Di Wu}
  is a senior staff research scientist, a team leader at Samsung
AI Center Montreal and an Adjunct Professor at McGill University since
2019. He did postdoctoral research at Montreal MILA and Stanford University.
He received his Ph.D. and M.Sc. from McGill University in 2018 and
Peking University in 2013, respectively. Di also holds Bachelor's
degrees in microelectronics and economics. His research interests
mainly lie in reinforcement learning, transfer learning, meta-Learning,
and multitask Learning. He is also interested in leveraging such algorithms
to improve real-world systems.
\end{IEEEbiography}

\end{document}